\documentclass{article} 
\usepackage{iclr2027_conference,times}

\iclrfinalcopy
\usepackage{amsmath,amsfonts,bm}

\def\eqref#1{(\ref{#1})}

\def\1{\bm{1}}

\DeclareMathAlphabet{\mathsfit}{\encodingdefault}{\sfdefault}{m}{sl}
\SetMathAlphabet{\mathsfit}{bold}{\encodingdefault}{\sfdefault}{bx}{n}

\newcommand{\E}{\mathbb{E}}

\usepackage{url}

\usepackage{amsmath,amssymb,amsthm,mathtools}
\usepackage{placeins}

\usepackage{booktabs}
\usepackage{graphics}
\usepackage{multirow}
\usepackage{makecell}
\usepackage{ragged2e}
\usepackage{float}
\usepackage{caption}

\usepackage{booktabs}
\usepackage{multirow}
\usepackage[table]{xcolor}
\usepackage{graphicx}
\usepackage{wrapfig}

\usepackage{marvosym} 
\usepackage{fontawesome5}

\usepackage{enumitem}
\usepackage[colorlinks=true,linkcolor=inkblue,citecolor=inkblue,urlcolor=inkblue]{hyperref}
\definecolor{inkblue}{HTML}{24597A}
\definecolor{methodgreen}{HTML}{007A67}
\definecolor{methodwash}{HTML}{EDF6F3}

\newcommand{\method}{STEPQuant}

\newcommand{\diag}{\operatorname{diag}}
\newcommand{\tr}{\operatorname{tr}}

\newcommand{\gm}{\operatorname{GM}}

\newcommand{\norm}[1]{\left\lVert #1\right\rVert}
\newtheorem{proposition}{Proposition}

\newcommand{\lin}[1]{\textcolor{black}{#1}}
\newcommand{\yao}{\color{black}}
\newenvironment{hbenv}{\begingroup\color{black}}{\endgroup}
\newcommand{\hb}[1]{\textcolor{black}{#1}}

\graphicspath{{figures/dl/}}

\title{STEPQuant: When and Where Errors Matter \\ in Delta-Rule Recurrent State Quantization}

\author{
Bingchen Yao$^{*1}$,
Haobo Xu$^{*3}$,
Haokun Lin $^{\ddagger2,4}$\textsuperscript{\footnotesize\Letter},
Yichen Wu$^{5}$,
\\
\textbf{
Ziyu Guo$^{6}$, 
Renrui Zhang$^{6}$, 
Zhichao Lu$^{4}$, 
Zhenan Sun$^{2}$,
Ying Wei$^{1}$\textsuperscript{\footnotesize\Letter}
}
\vspace{0.2cm}
\\
$^1$ Zhejiang University \quad
$^2$ NLPR \& MAIS, Institute of Automation, CAS \quad
$^3$ Tsinghua University \\
$^4$ City University of Hong Kong \quad
$^5$ Harvard University \quad
$^6$ The Chinese University of Hong Kong \\
\vspace{0.2cm}
\normalsize  $^*$Equal Contribution\hspace{0.2cm} $^\ddagger$Project Leader\hspace{0.2cm} \textsuperscript{\footnotesize\Letter}Corresponding Author \\
\vspace{0.15cm}
\faGithub\ \textbf{Code:} 
{\url{https://github.com/Dreamer-Toby/STEPQuant}}\\
}

\begin{document}

\maketitle

\begin{abstract}
Linear attention replaces growing KV caches with fixed-size recurrent states, yet these persistent states can become a substantial memory bottleneck under concurrent serving. Directly quantizing recurrent states to low precision often leads to severe accuracy degradation, as quantization errors propagate through successive state updates. We discover that the impact of these errors 
depends on two complementary dimensions: \textit{temporally}, errors in long-lived memory can persist across many decoding steps; \textit{spatially}, errors in different key rows affect model outputs differently, while state magnitudes vary substantially along both rows and columns. Motivated by these observations, we propose STEPQuant, a spatial-temporal post-training quantization framework for Delta-rule recurrent states. STEPQuant allocates precision according to error magnitude and memory lifetime, and jointly fits key-row and value-column scales based on state distributions and key-row impact on output error. Experiments on Qwen3.8-27B and Kimi-Linear-48B-A3B-Instruct across both long- and short-generation benchmarks show that STEPQuant closely matches FP32-state accuracy under a nominal 6-bit budget and outperforms uniform INT8 in its 4-bit configuration. Integrated into SGLang with optimized GPU kernels, 6-bit STEPQuant achieves over 5$\times$ recurrent-state compression and reduces total serving memory by up to 68.7\%.
\end{abstract}

\section{Introduction}

\begin{figure}[!t]
 \noindent\makebox[\linewidth][l]{\includegraphics[width=1.\linewidth]{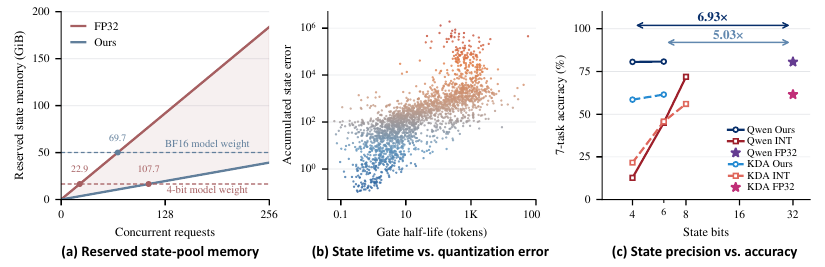}}
\caption{
\textbf{Why recurrent state needs structured compression.}
(a) Recurrent state memory grows with concurrent requests and can exceed model weight memory. STEPQuant substantially reduces memory cost at 6bit. 
(b) Heads with longer gate half-lives tend to accumulate larger state errors under uniform INT6 quantization across 2304 Qwen heads.
(c) Mean accuracy across seven tasks for Qwen (solid) and KDA (dashed). STEPQuant achieves $6.93\times$ and $5.03\times$ compression of Qwen recurrent states at 4 and 6 bits, respectively, with negligible degradation in mean accuracy.
}
 \label{fig:state_capacity}
\end{figure}

\begin{hbenv}
Unlike conventional softmax attention, which maintains a KV cache that grows with sequence length, linear attention summarizes past tokens into a fixed-size recurrent state~\citep{yang2023gla}. 
Recent hybrid models, including Qwen3.8-27B~\citep{qwen2026qwen38} and Kimi-Linear-48B-A3B-Instruct~\citep{kimi2025linear}, combine gated Delta-rule recurrent memory~\citep{yang2025gated} with standard attention 
\lin{to balance efficiency and performance. 
Despite its fixed size across context length, the recurrent state introduces a different memory bottleneck during serving.
Each concurrent request requires a separate persistent state, while serving systems may reserve additional state slots for caching and scheduling.
As a result, the memory footprint of the state pool \textit{grows with concurrency even for short contexts}.
In official SGLang deployment, the FP32 state pool of Qwen exceeds the memory footprint of its BF16 weights at 70 supported concurrent requests, as shown in Fig.~\ref{fig:state_capacity}(a).
This growing memory overhead motivates efficient compression of recurrent states.
}

However, directly applying uniform quantization to recurrent states severely degrades model performance. 
\lin{
Each Delta update operates on an approximate state and produces a newly quantized one, allowing quantization errors to propagate through subsequent decoding steps. 
}
Fig.~\ref{fig:state_capacity}(c) shows that mean accuracy drops sharply at 4 and 6 bits on both models, and a clear gap remains even at 8 bits.

\lin{
To understand this failure, we investigate state quantization error from two complementary dimensions: \textit{when} and \textit{where} the error accumulates.
First, \textit{temporally}, recurrent states are updated and requantized at every decoding step, causing quantization errors to accumulate over time.
Each update carries forward existing error while introducing new quantization error, and stronger gate retention allows these errors to persist longer.
Our preliminary experiments discover that longer-lived state units tend to exhibit larger accumulated errors (Fig.~\ref{fig:state_capacity}(b)).
Motivated by this observation, we introduce Lifetime-aware Bit Allocation, a mixed-precision quantization method that jointly considers the quantization error of each unit at different bit widths and its persistence over time.
Under a fixed memory budget, it allocates higher precision to units with larger and longer-lived errors.
}

\lin{
Second, \textit{spatially}, the effect of quantization error also depends on where it occurs in the recurrent state.
The spatial structure matters in two ways: errors of similar magnitude in different key rows can have different impacts on the model output, while state magnitudes vary substantially across both rows and columns (Fig.~\ref{fig:read_sensitivity}). 
Scaling along a single axis cannot accommodate both patterns as the state evolves during decoding.
We therefore propose Key-Row-Aware Dual-axis Fitting to reduce spatial quantization error. 
Specifically, we assign separate scales to key rows and value columns. The row scales account for both the current magnitude of each row and its measured impact on output error, while the column scales are fitted with greater weight on rows where errors matter more.
}

\lin{
Together, these temporal and spatial designs form \textbf{S}patial-\textbf{TE}m\textbf{P}oral \textbf{Quant}ization (STEPQuant), which compresses recurrent states throughout decoding.
We evaluate STEPQuant on two strong hybrid models, Qwen3.8-27B and Kimi-Linear-48B-A3B-Instruct, across seven long-generation and six short-generation tasks.
Under a nominal 6-bit budget, STEPQuant closely matches FP32-state accuracy on both models. 
Even with a 4-bit budget, STEPQuant outperforms uniform INT8 on both models, nearly matching FP32-state performance on Qwen while maintaining a modest gap on Kimi.
With optimized kernels integrated into SGLang, 6-bit STEPQuant compresses recurrent-state memory by 5.03$\times$ and 5.08$\times$, reducing total memory consumption, including model weights, by 68.7\% and 53.7\% on Qwen and Kimi, respectively.
Our contributions are summarized as follows: 
}
\begin{itemize}
[leftmargin=3mm, itemsep=-0.1mm, topsep=-0.1mm]
\item \textbf{Error analysis.} We analyze how quantization errors propagate through gated Delta-rule updates temporally and spatially, showing that both state lifetime and error location affect their impact.
\item \textbf{Quantization method.} 
\lin{
We propose STEPQuant, which combines lifetime-aware bit allocation with key-row-aware dual-axis fitting. Under a nominal 6-bit budget, STEPQuant closely matches FP32-state accuracy, while its 4-bit configuration outperforms uniform INT8 on both models.
}
\item \textbf{Serving implementation.} 
\lin{
We integrate STEPQuant into SGLang with optimized GPU kernels, achieving over 5$\times$ recurrent-state compression and up to 2.91$\times$ faster state updates.
}
\end{itemize}

\end{hbenv}

\section{Preliminaries}
\label{sec:preliminaries}

\begin{hbenv}
\subsection{Gated Delta-Rule Linear Attention}
\label{sec:delta_attention}

Unlike softmax attention, which maintains a KV cache that grows with
sequence length, recurrent linear attention summarizes past tokens in
a fixed-size state matrix.
\lin{
For a single head in a given layer, we omit the layer and head indices and denote the state after token $t$ by $S_t \in \mathbb{R}^{d_k \times d_v}$, with query $q_t \in \mathbb{R}^{d_k}$, key $k_t \in \mathbb{R}^{d_k}$, and value $v_t \in \mathbb{R}^{d_v}$.
}
Gated DeltaNet (GDN)~\citep{yang2025gated} and
Kimi Delta Attention (KDA)~\citep{kimi2025linear} update this state through
a gated Delta rule:
\begin{equation}
\label{eq:delta_update}
    S_t
    =
    D_t S_{t-1}
    +
    \beta_t k_t
    \bigl(v_t^\top - k_t^\top D_t S_{t-1}\bigr)=(I - \beta_t k_t k_t^\top)D_tS_{t-1}+\beta_t k_t v_t^\top,
    \qquad
    y_t = S_t^\top q_t,
\end{equation}
where $D_t$ controls memory retention and
$\beta_t \in [0,1]$ controls the write strength.
The update first applies the retention gate $D_t$ to the previous state, yielding the retained state $D_tS_{t-1}$. 
\lin{
The current key $k_t$ retrieves $k_t^\top D_tS_{t-1}$ from the retained state.
The residual $v_t^\top-k_t^\top D_tS_{t-1}$ is used to update the state along the key direction $k_t$. 
The head output (readout) is then computed as $y_t=S_t^\top q_t$.
}

The two architectures differ in their retention gate $D_t$.
GDN uses a scalar gate per head, $D_t = \alpha_t I$,
while KDA uses channel-wise gates
$D_t = \text{diag}(d_{t,1}, \cdots, d_{t,d_k})$.
Both updates can be written as
\begin{equation}
\label{eq:delta_affine}
    S_t = A_t S_{t-1} + B_t,
    \qquad \text{where}\qquad
    A_t = (I - \beta_t k_t k_t^\top)D_t,
    \qquad
    B_t = \beta_t k_t v_t^\top.
\end{equation}
Here, $A_t\in\mathbb{R}^{d_k\times d_k}$ transports and selectively modifies the previous memory,
while $B_t\in\mathbb{R}^{d_k\times d_v}$ introduces the new value.
The state contains $d_k\cdot d_v$ elements per head, independent of
sequence length, but must persist across decoding steps for each
active request.

\subsection{Recurrent-State Quantization}
\label{sec:state_quantization}

Quantization maps floating-point values to a finite set of discrete
levels, reducing storage requirements.
In symmetric uniform quantization, each value $x$ is encoded as a $b$-bit integer with a positive scale $s$ shared within a quantization group. The reconstructed value is
\begin{equation}
\label{eq:uniform_quantization}
    \mathcal{Q}_{b,s}(x)
    =
    s \cdot
    \operatorname{clip}
    \left(
        \operatorname{round}\left(\frac{x}{s}\right),
        -q_b, q_b
    \right),
    \qquad
    q_b = 2^{b-1}-1,
\end{equation}
\lin{
We focus on quantizing persistent recurrent states and also evaluate the combination with weight quantization.
} 
Let $S_t$ denote the full-precision reference state obtained by recursively applying Eq.~\eqref{eq:delta_affine} without state quantization.
\lin{
At each decoding step, the model updates the reconstructed state in floating point and computes the output. The updated state is then quantized for storage:
}
\begin{equation}
\label{eq:compressed_recurrence}
    X_t = A_t \hat{S}_{t-1} + B_t,
    \qquad
    \hat{y}_t = X_t^\top q_t,
    \qquad
    \hat{S}_t = \mathcal{Q}_t(X_t),
\end{equation}
where $\mathcal{Q}_t$ quantizes $X_t$ and returns its dequantized approximation $\hat S_t$.
\end{hbenv}
\section{Temporal Dimension: Lifetime-aware Bit Allocation}
\label{sec:analysis}
\subsection{Lifetime-Dependent Error Accumulation}
\label{subsec:lifetime}

\hb{
From Eq.~\ref{eq:compressed_recurrence}, quantization error is recursively fed back through the state update. The following proposition characterizes how this error propagates across decoding steps.
Proof is in Appendix~\ref{app:proofs}.
}
\begin{proposition}[Conditional error propagation]
\label{prop:error}
For identical inputs and gates, \hb{let
$E_t=\widehat S_t-S_t$ denote the accumulated error and
$\varepsilon_t=\mathcal Q_t(X_t)-X_t$ the quantization error added at step $t$. Then}
\begin{equation}
 E_t=(I - \beta_t k_t k_t^\top)D_tE_{t-1}+\varepsilon_t=A_tE_{t-1}+\varepsilon_t,\qquad
 \widehat y_t-y_t=E_{t-1}^\top A_t^\top q_t.\label{eq:error}
\end{equation}
If $\|k_t\|_2\leq1$, $0\leq\beta_t\leq1$, and $0\preceq D_t\preceq I$,
then $\|A_t\|_2\leq\|D_t\|_2\leq1$.
\end{proposition}

\begin{hbenv}
According to Proposition~1, previously accumulated error is propagated through $A_t$, while quantization at step $t$ introduces a new error.
\lin{
The transition $A_t$ reduces existing error through two mechanisms. 
The retention gate $D_t$ first attenuates the error carried over from the previous step. 
The Delta update then further reduces its component along the current key $k_t$ through the factor $(I-\beta_tk_tk_t^\top)$, while leaving components orthogonal to $k_t$ unchanged.
We verify this effect experimentally in Appendix~\ref{app:oracle}.
Thus, errors in directions rarely aligned with subsequent keys depend mainly on gate decay. 
When retention is close to one, these errors can persist for many decoding steps.
Together, \textbf{these results identify memory lifetime as a key predictor of recurrent-state quantization risk.}
}

This analysis indicates that longer-lived memory should exhibit larger
accumulated quantization error.  
To examine this assumption, we evaluate all recurrent heads of
Qwen3.8-27B under uniform INT6 state quantization on C4.
Fig.~\ref{fig:state_capacity}(b) shows that heads with longer gate
half-lives exhibit larger accumulated state error, with Spearman's
$\rho_S \approx 0.80$
(see Appendix~\ref{app:lifetime_state} for more details).
This indicates that \textbf{recurrent heads with longer lives tend to result in larger quantization error.}
\end{hbenv}

\begin{figure}[!ht]
 \centering\includegraphics[width=\linewidth]{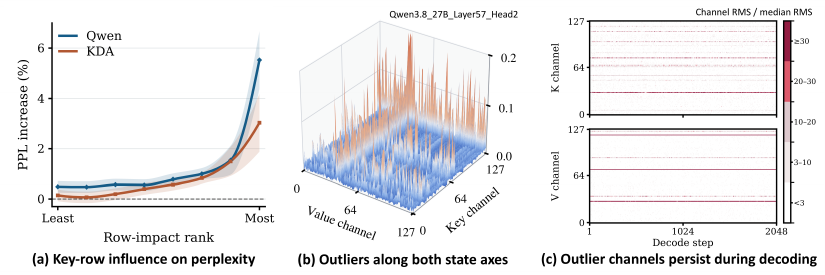}
 \caption{
\textbf{Spatial structure of recurrent states.}
(a) Key rows are ranked by readout impact and divided into eight groups. Quantizing one group at a time to INT4 generally causes larger perplexity increases for higher-impact groups.
(b) A representative Qwen state exhibits obvious outliers in both key rows and value columns.
(c) Channel RMS relative to the median across 2048 decoding steps. The outliers remain prominent throughout decoding.
 }
 \label{fig:read_sensitivity}
\end{figure}

\begin{hbenv}
\subsection{Lifetime-aware Bit Allocation}

\label{sec:lmsq}

\lin{
Mixed-precision quantization is widely used in LLMs to preserve accuracy by assigning more bits to sensitive components~\citep{dettmers2022gpt3,dettmers2023spqr}. 
For recurrent states, bit allocation should additionally account for memory lifetime, as quantization errors can persist across subsequent updates.
Consequently, we propose Lifetime-aware Bit Allocation, 
which assigns precision under a fixed bit budget based on the persistence of quantization errors.
}

\lin{
Lifetime-aware Bit Allocation operates on recurrent-state units: each unit $u$ is an entire head in Qwen and a key row in KDA.
We estimate $d_u(b)$, the reconstruction distortion of unit $u$ quantized to $b$ bits, and its mean log retention $\ell_u$
where the expectation is taken over calibration tokens.
Here, $r_{t,u}$ is the retention gate for unit $u$: $r_{t,u}=\alpha_t$ for GDN and $r_{t,u}=d_{t,i}$ for key row $i$ in KDA.
A larger $\ell_u$ indicates stronger retention and a longer memory lifetime.
Considering gate retention alone, the error retention factor after $j$ updates is approximated by
}
\begin{equation}
\prod_{s=1}^{j} r_{t+s,u}
\approx \exp(j\ell_u).
\end{equation}
\lin{
The squared error therefore decays by a factor of approximately $\exp(2j\ell_u)$, giving the lifetime weight over $H$ steps:
}
\begin{equation}
L_u=\sum_{j=0}^{H-1}\exp(2j\ell_u).
\end{equation}
\lin{
Let $n_u$ denote the number of state elements in unit $u$. Given an average bit budget $\bar b$, we select bit widths from the candidate set $\mathcal B_{\bar b}$ to minimize the lifetime-weighted reconstruction distortion:
}
\begin{equation}
\min_{{b_u\in\mathcal B_{\bar b}}}
\sum_u L_u d_u(b_u)
\quad\text{s.t.}\quad
\sum_u n_u b_u
\leq
\bar b\sum_u n_u.
\label{eq:allocation}
\end{equation}
\lin{
This objective accounts for both \textit{the magnitude of quantization error} and \textit{how long it persists through subsequent updates}.
Nevertheless, a small number of units remain difficult to quantize even with mixed-precision allocation. We therefore retain the highest-risk units in FP16 as sparse pivots.
}

\lin{
In summary, Lifetime-aware Bit Allocation assigns higher precision to units with larger quantization errors and longer memory lifetimes. The pivot selection and bit allocation are determined during offline calibration and fixed across requests. Further details are provided in Appendix~\ref{app:calibration} and~\ref{app:capacity}.
}

\end{hbenv}

\section{Spatial Dimension: Key-Row-Aware Dual-Axis Fitting}

\begin{hbenv}

\subsection{Key-row Impact on Readout Error}
\label{sec:key-row}
The temporal analysis above characterizes how quantization errors propagate across recurrent updates. 
\lin{
We now examine their spatial distribution within the state matrix: errors of equal magnitude in different key rows can affect the readout differently. From Eq.~\eqref{eq:error}, the readout error is
}
\begin{equation}
\label{eq:4.2}
\Delta y_t = \hat y_t-y_t
=
E_{t-1}^{\top}A_t^{\top}q_t
=
\sum_i (A_t^{\top}q_t)_i E_{t-1,i,:}^{\top},
\end{equation}
\lin{
We define $g_t=A_t^\top q_t\in\mathbb{R}^{d_k}$, where $g_{t,i}$ weights the contribution of the error in key row $i$, $E_{t-1,i,:}$, to the current readout error.
For errors of equal norm confined to individual key rows, a larger $g_{t,i}^2$ implies a larger squared readout error.
To account for variation across decoding steps, we define the row-impact score as
}
\begin{equation} \omega_i=\mathbb{E}_{\mathrm{cal}}[g_{t,i}^2]. \end{equation}
To validate this row-impact score, we rank key rows within each head by \(\omega_i\) and divide them into eight groups. We quantize one group at a time to INT4 while keeping the remaining rows in full precision, and measure the resulting increase in perplexity. As shown in Fig.~\ref{fig:read_sensitivity}(a), groups with larger \(\omega_i\) generally produce greater PPL degradation in both models.
\lin{
This observation suggests \textbf{using the row-impact score to guide quantization scale selection}, as described in Sec.~\ref{sec:fitting}. 
}
Notably, squaring Eq.~\eqref{eq:4.2} introduces cross-row interaction terms in the readout error, which are not explicitly captured by our row-impact score. However, their net contribution is at most 3.1\% at $H=1024$ under STEPQuant, as discussed in Appendix~\ref{a.2}.

\subsection{Two-Axis State Geometry}
\label{sec:two-axis}

\lin{
Beyond differences in how key-row errors affect the readout, recurrent states exhibit another spatial property: large-magnitude outliers occur along both key rows and value columns.
}
Unlike conventional LLM quantization settings that often target outliers along a single dominant axis (e.g., channels or tokens)~\citep{xiao2023smoothquant,shao2024omniquant}, we discover that the recurrent state matrix exhibits large-magnitude structures along both key rows and value columns, as shown in Fig.~\ref{fig:read_sensitivity}(b). 
We further examine how these magnitude patterns evolve throughout decoding. 
Fig.~\ref{fig:read_sensitivity}(c) shows that pronounced magnitude differences persist along both axes, with a small subset of rows and columns consistently exhibiting substantially larger magnitudes. Quantitative results show that the maximum-to-median RMS contrasts along the key-row and value-column axes are $10.3\times$ and $19.4\times$, respectively. 
Both exceed $3\times$ in 98.6\% of sampled states. 
Together, these observations motivate \textbf{dual-axis scaling to accommodate the magnitude distributions of recurrent states}.

\subsection{Key-Row-Aware Dual-axis Fitting}
\label{sec:fitting}
\lin{
Sec.~\ref{sec:key-row} and~\ref{sec:two-axis} identify two spatial properties of recurrent states: key rows differ in their impact on readout error, and state magnitudes vary substantially along both axes. 
To account for both, we propose Key-Row-Aware Dual-Axis Fitting, which combines calibrated row-impact scores with separate row and column scales. 
Specifically, we represent the updated state $X=X_t$ as:
}

\begin{equation} \hat X_{ij}=r_i c_j z_{ij}, \end{equation}

where \(r_i>0\) and \(c_j>0\) are the scale factors for key row \(i\) and value column \(j\), respectively, and \(z_{ij}\) is the low-bit integer. We incorporate the key-row impact factor into $r_i$, and the two factors allow the
quantization independently to vary along the two
state axes.

\paragraph{Row factors.}
The row factor $r_i$ should account for both (i) the current magnitude of a row
and (ii) the impact of key row on readout error (Sec.~\ref{sec:key-row}). We first estimate the magnitude
of key row $i$ as
\begin{equation}
m_i=\frac{1}{d_v}\sum_j |X_{ij}|.
\end{equation}
Rows with larger $m_i$ require a wider quantization range, so the row
factor should increase with $m_i$.
In addition, key rows with larger
impact scores $\omega_i$ are more vulnerable to quantization.
Inspired by fractional-power smoothing in prior quantization methods~\citep{xiao2023smoothquant},
we consider both effects using square-root scaling and a dimension-normalized $w_i$\footnote{We temper the dynamic range and normalize within each head as
$ w_i= \omega_i^{\gamma/2}/\operatorname{GM}_k(\omega_k^{\gamma/2})$, $\gamma=0.25, $
so that the geometric mean of $w_i$ is one while preserving relative row importance, following~\citep{box1964analysis}.}:
\begin{equation}
r_i = m_i^{1/2} w_i^{-1/2}.
\end{equation}
Thus, larger-magnitude rows receive a wider range, while higher-impact
rows receive finer quantization resolution.
\paragraph{Column factors.} Given the row factors $\{r_i\}$, we fit the column scales $\{c_j\}$ by minimizing the
impact-weighted reconstruction error:
\begin{equation}
\min_{\{c_j>0\}}
\sum_{i,j}
w_i^2
\left(X_{ij}-r_i c_j z_{ij}\right)^2 .
\end{equation}
This objective captures the remaining variation along value columns,
while assigning larger penalties to reconstruction errors on
high-impact key rows.

\vspace{-10pt}
\paragraph{Quantization.} 
\lin{
Given the fitted scales, we quantize each scaled entry $X_{ij}/(r_i c_j)$ to the nearest representable level at its assigned precision $b_i$. The reconstructed state is then $\widehat X_{ij}=r_i c_j z_{ij}$.
Implementation details of scale fitting and quantized-state storage are provided in Appendix~\ref{app:algorithm}.
}

\end{hbenv}

\subsection{Kernel Implementation in SGLang}
\label{sec:kernel}

\lin{
We implement STEPQuant as packed-state kernels integrated with SGLang's recurrent-state pool. 
Lifetime-Aware Bit Allocation and FP16 pivot selection are performed offline, adding no per-token allocation overhead and fixing the packed layout across requests. 
During decoding, we implement a kernel to fuse tilewise state reconstruction, the Delta update, and the current readout. 
This avoids a separate pass over the full state for reconstruction and reduces memory traffic.
Once the head output is available, later layers continue processing the token while Key-Row-Aware Dual-Axis Fitting and packed writeback run on a separate CUDA stream. 
This overlaps scale updates with model computation and keeps the state compressed between tokens (see more details in Appendix~\ref{app:backends}).
}

{\yao
\section{Experiments}
\label{sec:experiments}

\subsection{Setup}
\label{sec:protocol}
\noindent\textbf{Models and hardware.}
We evaluate Qwen3.8-27B~\citep{qwen2026qwen38} and Kimi-Linear-48B-A3B-Instruct~\citep{kimi2025linear} with BF16 and 4-bit AWQ-quantized weights~\citep{lin2024awq}.
To further assess generalization across model sizes and architectures, we additionally evaluate Qwen3.5-9B~\citep{qwen3.5} and Qwen3.6-27B~\citep{qwen3.6-27b} as dense GDN models, and Qwen3.6-35B-A3B~\citep{qwen3.6-35b-a3b} as a GDN-based MoE model.
All experiments are conducted on SGLang~\citep{zheng2024sglang} using NVIDIA A800 GPUs.
We use SGLang’s default FP32 SSM-state precision as the full-precision baseline, alongside symmetric rowwise-absmax INT4/6/8 baselines.
\lin{
STEPQuant uses the fused state kernels described in Sec.~\ref{sec:kernel}.
}

\noindent\textbf{Calibration.}
For all measured benchmarks,
we use 32 WikiText-2 \citep{merity2016pointer} training segments with 2048
tokens each to determine bit allocation, FP16 pivots, and row-impact scores.

\noindent\textbf{Long-generation reasoning.}
\lin{
We compare quantized models on seven reasoning benchmarks: LiveCodeBench v6~\citep{jain2025livecodebench}, EvalPlus~\citep{liu2023your}, AIME 2026~\citep{dekoninck2026beyond}, MATH-500~\citep{lightman2024let}, HMMT February 2026~\citep{dekoninck2026beyond}, GPQA Diamond~\citep{rein2023gpqa}, and IFBench~\citep{pyatkin2026generalizing}.
We generate 5, 5, 64, 4, 64, 8, and 4 samples per question, respectively, using temperature $T=1.0$, top-$k=20$, top-$p=0.95$, and a maximum of 65536 generated tokens per sample.
}

\noindent\textbf{Short generation.}
We also report results on six language understanding tasks: MMLU~\citep{hendrycks2020measuring}, ARC-C~\citep{clark2018think}, OpenBookQA~\citep{mihaylov2018can}, HellaSwag~\citep{zellers2019hellaswag}, WinoGrande~\citep{sakaguchi2021winogrande}, and LAMBADA~\citep{paperno2016lambada}.
We generate one answer per example with greedy decoding ($T=0$) and score the generated answers rather than candidate likelihoods.
This aligns the autoregressive decoding setting targeted by STEPQuant.

\noindent\textbf{Long-context evaluation.}
We further evaluate STEPQuant on RULER NIAH~\citep{hsieh2024ruler} at 32K and 128K context lengths and LongBench-E~\citep{bai2024longbench}, including 13 tasks.

\begin{table}[t]
 \centering\small\setlength{\tabcolsep}{3pt}
 \caption{\textbf{Long-generation reasoning accuracy with BF16 weights (\%).}
 }
 \resizebox{0.95\linewidth}{!}{
 \begin{tabular}{@{}lrrrrrrrrr@{}}
\toprule
State & Bits & LCB v6 & EvalPlus & AIME 26 & MATH-500 & HMMT & GPQA-D & IFBench & Avg. \\
\midrule
\multicolumn{10}{@{}l}{\textit{Qwen3.8-27B}} \\
\addlinespace[2pt]
FP32 & 32 & 85.31 & 84.87 & 87.71 & 97.60 & 74.24 & 80.81 & 53.67 & 80.60 \\
INT8 & 8 & 72.23 & 80.26 & 78.54 & 97.00 & 58.71 & 75.25 & 41.00 & 71.86 \\
INT6 & 6 & 30.43 & 69.00 & 33.96 & 87.40 & 16.67 & 51.52 & 26.33 & 45.04 \\
INT4 & 4 & 7.87 & 31.55 & 0.00 & 34.00 & 0.00 & 4.04 & 11.67 & 12.73 \\
DAMP~\citep{zhang2026damp} & 9.88 & 84.45 & 84.50 & 86.88 & 97.70 & 73.11 & 85.86 & 53.67 & 80.88 \\
DAMP~\citep{zhang2026damp} & 6.38 & 13.46 & 48.06 & 19.79 & 77.90 & 17.05 & 42.93 & 18.33 & 33.93 \\
\rowcolor{methodwash}
\textbf{STEPQuant@6bit} & \textbf{6.36} & \textbf{85.42} & \textbf{85.54} & \textbf{87.24} & \textbf{97.35} & \textbf{73.30} & \textbf{80.87} & \textbf{54.42} & \textbf{80.59} \\
\rowcolor{methodwash}
\textbf{STEPQuant@4bit} & \textbf{4.62} & \textbf{86.35} & \textbf{84.87} & \textbf{86.25} & \textbf{97.00} & \textbf{73.86} & \textbf{81.57} & \textbf{53.67} & \textbf{80.51} \\
\midrule
\multicolumn{10}{@{}l}{\textit{Kimi-Linear-48B-A3B-Instruct}} \\
\addlinespace[2pt]
FP32 & 32 & 54.52 & 76.75 & 68.33 & 93.25 & 45.08 & 69.70 & 23.00 & 61.52 \\
INT8 & 8 & 52.61 & 75.46 & 51.88 & 92.45 & 35.61 & 62.88 & 21.25 & 56.02 \\
INT6 & 6 & 42.37 & 71.96 & 22.71 & 88.00 & 17.42 & 57.83 & 19.58 & 45.70 \\
INT4 & 4 & 28.82 & 46.86 & 0.00 & 45.80 & 0.00 & 14.58 & 15.33 & 21.63 \\
DAMP~\citep{zhang2026damp} & 9.88 & 51.94 & 76.94 & 62.08 & 93.40 & 41.86 & 67.17 & 22.33 & 59.39 \\
DAMP~\citep{zhang2026damp} & 6.38 & 18.01 & 53.04 & 31.04 & 81.40 & 19.70 & 52.53 & 16.00 & 38.82 \\
\rowcolor{methodwash}
\textbf{STEPQuant@6bit} & \textbf{6.30} & \textbf{54.86} & \textbf{76.57} & \textbf{67.71} & \textbf{93.60} & \textbf{46.21} & \textbf{68.43} & \textbf{22.92} & \textbf{61.47} \\
\rowcolor{methodwash}
\textbf{STEPQuant@4bit} & \textbf{4.30} & \textbf{53.29} & \textbf{74.54} & \textbf{59.17} & \textbf{92.60} & \textbf{43.56} & \textbf{64.14} & \textbf{22.33} & \textbf{58.52} \\
\bottomrule
\end{tabular}

 }
 \label{tab:reasoning}
\par\vspace{8pt}
 \setlength{\tabcolsep}{3.5pt}
 \caption{\textbf{Short-generation accuracy with BF16 weights (\%).}
 }
 \resizebox{0.95\linewidth}{!}{
 \begin{tabular}{@{}lrrrrrrrr@{}}
\toprule
State & Bits & MMLU & ARC-C & OpenBookQA & HellaSwag & WinoGrande & LAMBADA & Avg. \\
\midrule
\multicolumn{9}{@{}l}{\textit{Qwen3.8-27B}} \\
\addlinespace[2pt]
FP32 & 32 & 82.25 & 96.93 & 95.40 & 93.15 & 90.06 & 68.91 & 87.78 \\
INT8 & 8 & 82.19 & 96.16 & 94.60 & 92.00 & 87.69 & 64.89 & 86.25 \\
INT6 & 6 & 78.86 & 94.28 & 92.80 & 87.72 & 79.79 & 61.50 & 82.49 \\
INT4 & 4 & 78.80 & 73.63 & 77.60 & 60.51 & 46.57 & 57.31 & 65.74 \\
\rowcolor{methodwash}
\textbf{STEPQuant@6bit} & \textbf{6.36} & \textbf{82.23} & \textbf{96.67} & \textbf{95.80} & \textbf{93.01} & \textbf{89.19} & \textbf{68.35} & \textbf{87.54} \\
\rowcolor{methodwash}
\textbf{STEPQuant@4bit} & \textbf{4.62} & \textbf{82.07} & \textbf{96.84} & \textbf{95.80} & \textbf{93.08} & \textbf{89.50} & \textbf{68.50} & \textbf{87.63} \\
\midrule
\multicolumn{9}{@{}l}{\textit{Kimi-Linear-48B-A3B-Instruct}} \\
\addlinespace[2pt]
FP32 & 32 & 71.21 & 91.64 & 88.60 & 67.42 & 55.88 & 35.40 & 68.36 \\
INT8 & 8 & 70.67 & 91.98 & 88.40 & 66.78 & 55.80 & 33.17 & 67.80 \\
INT6 & 6 & 67.73 & 87.54 & 83.40 & 61.77 & 41.28 & 26.37 & 61.35 \\
INT4 & 4 & 57.36 & 69.37 & 72.80 & 19.88 & 17.36 & 16.67 & 42.24 \\
\rowcolor{methodwash}
\textbf{STEPQuant@6bit} & \textbf{6.30} & \textbf{71.21} & \textbf{90.61} & \textbf{89.40} & \textbf{68.77} & \textbf{57.70} & \textbf{35.20} & \textbf{68.82} \\
\rowcolor{methodwash}
\textbf{STEPQuant@4bit} & \textbf{4.30} & \textbf{71.08} & \textbf{90.96} & \textbf{88.80} & \textbf{67.95} & \textbf{54.93} & \textbf{34.93} & \textbf{68.11} \\
\bottomrule
\end{tabular}

 }
 \label{tab:short}
\end{table}

\subsection{Main Results}
\label{sec:results}

\noindent\textbf{Accuracy across long and short tasks.}
\label{sec:reasoning}
\lin{
For long reasoning tasks, Table~\ref{tab:reasoning} shows that 6-bit STEPQuant closely matches the FP32-state baseline on both Qwen (GDN) and Kimi (KDA).
STEPQuant achieves mean accuracies of 80.59\% on Qwen and 61.47\% on Kimi, whereas uniform INT6 reaches only 45.04\% and 45.70\%, respectively.
At 4 bits, uniform INT4 suffers substantial performance degradation, while STEPQuant achieves competitive performance across the seven tasks.
The advantage also holds on the short-generation benchmarks in Table~\ref{tab:short}.
Here, 4-bit STEPQuant trails FP32 by only 0.15 points on Qwen and 0.25 points on Kimi, whereas uniform INT4 degrades sharply and can fail to produce valid answers even on short tasks.
We further include a controlled comparison with DAMP under the same models and evaluation protocol. At 6.4 bits, STEPQuant retains near-FP32 accuracy, whereas DAMP degrades substantially. DAMP requires about 9.9 bits to achieve comparable accuracy, demonstrating that STEPQuant provides a markedly better accuracy-storage trade-off (see Appendix~\ref{app:damp_comparison}).
}

\begin{table}[!t]
 \centering\small\setlength{\tabcolsep}{2.5pt}
 \caption{\textbf{Reasoning performance of STEPQuant with 4-bit AWQ-quantized weights.}
 }
 \resizebox{0.95\linewidth}{!}{
 \begin{tabular}{@{}lrrrrrrrr@{}}
\toprule
State & LCB v6 & EvalPlus & AIME 26 & MATH-500 & HMMT & GPQA-D & IFBench & Avg. \\
\midrule
\multicolumn{9}{@{}l}{\textit{Qwen3.8-27B}} \\
\addlinespace[2pt]
FP32 & 85.23 & 85.06 & 82.92 & 97.20 & 71.78 & 81.06 & 52.00 & 79.32 \\
\rowcolor{methodwash}
\textbf{STEPQuant@6bit} & \textbf{85.33} & \textbf{85.42} & \textbf{82.76} & \textbf{97.25} & \textbf{71.02} & \textbf{80.93} & \textbf{52.17} & \textbf{79.27} \\
\rowcolor{methodwash}
\textbf{STEPQuant@4bit} & \textbf{83.98} & \textbf{84.69} & \textbf{85.42} & \textbf{96.40} & \textbf{70.83} & \textbf{81.06} & \textbf{50.50} & \textbf{78.98} \\
\midrule
\multicolumn{9}{@{}l}{\textit{Kimi-Linear-48B-A3B-Instruct}} \\
\addlinespace[2pt]
FP32 & 52.80 & 74.72 & 59.79 & 93.80 & 41.86 & 66.92 & 22.75 & 58.95 \\
\rowcolor{methodwash}
\textbf{STEPQuant@6bit} & \textbf{52.74} & \textbf{74.94} & \textbf{59.38} & \textbf{93.75} & \textbf{41.10} & \textbf{66.16} & \textbf{22.25} & \textbf{58.62} \\
\rowcolor{methodwash}
\textbf{STEPQuant@4bit} & \textbf{51.94} & \textbf{74.17} & \textbf{51.88} & \textbf{92.20} & \textbf{40.91} & \textbf{63.38} & \textbf{22.17} & \textbf{56.66} \\
\bottomrule
\end{tabular}

 }
 \label{tab:w4}
\end{table}

\begin{table*}[t]
\centering
\small
\setlength{\tabcolsep}{4.5pt}
\renewcommand{\arraystretch}{0.95}
\caption{\textbf{Generalization across different GDN model sizes and architectures.}}
\label{tab:gdn_generalization}
\resizebox{0.95\linewidth}{!}{
\begin{tabular}{@{}lrrrrrrrr@{}}
\toprule
State & LCB v6 & EvalPlus & AIME 26 & MATH-500 & HMMT & GPQA-D & IFBench & Avg. \\
\midrule
\multicolumn{9}{@{}l}{\textit{Qwen3.6-35B-A3B (GDN MoE)}} \\
\addlinespace[2pt]
FP32 & 68.06 & 81.55 & 77.08 & 98.80 & 52.84 & 76.77 & 36.00 & 70.16 \\
INT8 & 30.71 & 69.74 & 5.42 & 75.40 & 3.60 & 37.88 & 32.67 & 36.49 \\
INT6 & 2.56 & 31.00 & 0.00 & 12.20 & 0.19 & 13.13 & 24.33 & 11.92 \\
INT4 & 0.28 & 13.10 & 0.00 & 3.80 & 0.00 & 15.66 & 18.00 & 7.26 \\
\rowcolor{methodwash}
\textbf{STEPQuant@6bit} & \textbf{68.91} & \textbf{81.92} & \textbf{78.54} & \textbf{97.80} & \textbf{57.95} & \textbf{76.77} & \textbf{35.33} & \textbf{71.03} \\
\rowcolor{methodwash}
\textbf{STEPQuant@4bit} & \textbf{50.24} & \textbf{78.78} & \textbf{33.33} & \textbf{91.00} & \textbf{22.92} & \textbf{67.68} & \textbf{34.67} & \textbf{54.09} \\
\midrule
\multicolumn{9}{@{}l}{\textit{Qwen3.5-9B (Dense GDN)}} \\
\addlinespace[2pt]
FP32 & 58.77 & 69.37 & 72.92 & 95.80 & 51.89 & 78.28 & 37.67 & 66.39 \\
INT8 & 40.76 & 66.97 & 38.54 & 93.00 & 22.35 & 75.25 & 34.67 & 53.08 \\
INT6 & 13.36 & 42.62 & 0.00 & 45.60 & 0.00 & 15.66 & 29.33 & 20.94 \\
INT4 & 0.00 & 7.20 & 0.00 & 5.80 & 0.00 & 0.00 & 23.67 & 5.24 \\
\rowcolor{methodwash}
\textbf{STEPQuant@6bit} & \textbf{57.63} & \textbf{70.11} & \textbf{74.17} & \textbf{96.60} & \textbf{52.65} & \textbf{81.31} & \textbf{34.00} & \textbf{66.64} \\
\rowcolor{methodwash}
\textbf{STEPQuant@4bit} & \textbf{53.84} & \textbf{69.93} & \textbf{59.38} & \textbf{95.80} & \textbf{37.50} & \textbf{78.28} & \textbf{33.33} & \textbf{61.15} \\
\midrule
\multicolumn{9}{@{}l}{\textit{Qwen3.6-27B (Dense GDN)}} \\
\addlinespace[2pt]
FP32 & 69.76 & 81.73 & 79.79 & 98.40 & 56.44 & 77.27 & 38.67 & 71.72 \\
INT8 & 63.41 & 78.60 & 70.62 & 99.00 & 52.84 & 73.74 & 41.67 & 68.55 \\
INT6 & 36.59 & 64.58 & 30.00 & 90.00 & 24.24 & 64.14 & 37.00 & 49.51 \\
INT4 & 13.55 & 49.45 & 0.00 & 37.00 & 0.00 & 25.25 & 34.33 & 22.80 \\
\rowcolor{methodwash}
\textbf{STEPQuant@6bit} & \textbf{71.18} & \textbf{82.47} & \textbf{80.42} & \textbf{98.20} & \textbf{59.47} & \textbf{77.27} & \textbf{40.00} & \textbf{72.72} \\
\rowcolor{methodwash}
\textbf{STEPQuant@4bit} & \textbf{69.57} & \textbf{81.55} & \textbf{79.58} & \textbf{97.20} & \textbf{57.01} & \textbf{76.52} & \textbf{40.00} & \textbf{71.63} \\
\bottomrule
\end{tabular}
}
\end{table*}

\noindent\textbf{Compatibility with W4 weights.}
\lin{
We further evaluate STEPQuant with 4-bit AWQ-quantized weights, a setting more representative of practical deployment.
From Table~\ref{tab:w4}, 6-bit STEPQuant achieves seven-task average accuracies of 79.27\% on Qwen and 58.62\% on Kimi, only 0.05 and 0.33 points below their corresponding FP32-state baselines, respectively.
At the lower 4-bit state budget, STEPQuant also maintains competitive performance.
These results show that STEPQuant remains effective when combined with weight quantization.
As weight memory decreases, recurrent states account for a larger fraction of the serving memory, making state compression increasingly important for memory-efficient deployment.
}

\noindent\textbf{Long-context evaluation.}
As shown in Table~\ref{tab:long_context}, 6-bit STEPQuant closely matches FP32 across different sequence lengths, while 4-bit STEPQuant also maintains comparable performance, indicating that our method remains robust as the context length increases.

\noindent\textbf{Generalization across GDN variants.}
From Table~\ref{tab:gdn_generalization}, 6-bit STEPQuant consistently matches FP32 accuracy across all three models.
4-bit STEPQuant also remains substantially more accurate than uniform quantization, although the MoE model is more sensitive at this lower bit width. These results demonstrate that STEPQuant generalizes across different GDN model sizes and architectures.

\subsection{Component ablation}
\label{sec:ablation}
\begin{hbenv}
We evaluate the components on AIME 2026, GPQA Diamond, and LiveCodeBench v6 with BF16 weights under nominal 4- and 6-bit budgets.
We also adapt the dual-axis state quantization (DSQ) component of Q-Mamba~\citep{tianqi2025q}, originally designed for Mamba models, as a baseline. To isolate the effect of spatial quantization, we further combine Q-Mamba with STEPQuant's bit allocation and FP16 pivots while retaining its original quantization method.
Spatial only retains uniform precision but applies key-row-aware dual-axis
fitting. Temporal only uses calibrated mixed-precision allocation and FP16
pivots without spatial fitting, while Temporal w/o pivots removes the pivots.
STEPQuant combines the spatial and temporal components. All quantization methods are evaluated under the same nominal bit budget (4 or 6 bits). Table~\ref{tab:ablation} presents the
Qwen results, with the corresponding Kimi results
provided in Appendix~\ref{app:ablation}.

\begin{table*}[t]
\centering
\caption{\textbf{Long-context evaluation at different sequence lengths on RULER NIAH and LongBench.}
LongBench-E contains 13 English tasks.}
\label{tab:long_context}
\resizebox{0.95\linewidth}{!}{
\begin{tabular}{@{}lcccccc@{}}
\toprule
State & RULER-32K & RULER-128K & LB-E 0-4K & LB-E 4-8K & LB-E 8K+ & LB-E Avg. \\
\midrule
\multicolumn{7}{@{}l}{\textit{Qwen3.8-27B}} \\
\addlinespace[2pt]
FP32 & 100.00 & 100.00 & 59.51 & 57.36 & 54.26 & 57.22 \\
\rowcolor{methodwash}
\textbf{STEPQuant@6bit} & \textbf{100.00} & \textbf{100.00} & \textbf{59.64} & \textbf{57.47} & \textbf{54.38} & \textbf{57.30} \\
\rowcolor{methodwash}
\textbf{STEPQuant@4bit} & \textbf{100.00} & \textbf{100.00} & \textbf{59.29} & \textbf{57.07} & \textbf{54.17} & \textbf{57.01} \\
\midrule
\multicolumn{7}{@{}l}{\textit{Kimi-Linear-48B-A3B-Instruct}} \\
\addlinespace[2pt]
FP32 & 99.50 & 99.53 & 54.93 & 50.23 & 48.98 & 51.63 \\
\rowcolor{methodwash}
\textbf{STEPQuant@6bit} & \textbf{99.56} & \textbf{99.41} & \textbf{54.70} & \textbf{49.85} & \textbf{49.25} & \textbf{51.39} \\
\rowcolor{methodwash}
\textbf{STEPQuant@4bit} & \textbf{99.62} & \textbf{99.50} & \textbf{53.93} & \textbf{50.00} & \textbf{48.67} & \textbf{51.07} \\
\bottomrule
\end{tabular}
}
\end{table*}

\begin{table}[!t]
 \centering
 \caption{\textbf{Component ablation on three long-reasoning tasks using Qwen3.8-27B BF16 weights.}
 }
 \begin{minipage}[t]{0.49\linewidth}
\centering\scriptsize\setlength{\tabcolsep}{4pt}
\renewcommand{\arraystretch}{0.95}
\textbf{Nominal 6-bit budget}\par\vspace{2pt}
\begin{tabular}{@{}lrrrr@{}}
\toprule
Variant & AIME & GPQA & LCB & Avg. \\
\midrule
FP32 & 87.71 & 80.81 & 85.31 & 84.61 \\
INT6 & 33.96 & 51.52 & 30.43 & 38.63 \\
Q-Mamba@6bit & 73.13 & 76.77 & 77.97 & 75.95 \\
Q-Mamba + alloc./pivots & 85.00 & 80.81 & 81.04 & 82.28 \\
Spatial only & 79.79 & 80.68 & 80.09 & 80.19 \\
Temporal w/o pivots & 60.42 & 71.97 & 69.00 & 67.13 \\
Temporal only & 74.58 & 79.67 & 72.64 & 75.63 \\
\rowcolor{methodwash}
STEPQuant@6bit & \textbf{87.24} & \textbf{80.87} & \textbf{85.42} & \textbf{84.51} \\
\bottomrule
\end{tabular}
\end{minipage}\hfill
\begin{minipage}[t]{0.49\linewidth}
\centering\scriptsize\setlength{\tabcolsep}{4pt}
\renewcommand{\arraystretch}{0.95}
\textbf{Nominal 4-bit budget}\par\vspace{2pt}
\begin{tabular}{@{}lrrrr@{}}
\toprule
Variant & AIME & GPQA & LCB & Avg. \\
\midrule
FP32 & 87.71 & 80.81 & 85.31 & 84.61 \\
INT4 & 0.00 & 4.04 & 7.87 & 3.97 \\
Q-Mamba@4bit & 0.00 & 6.06 & 16.87 & 7.64 \\
Q-Mamba + alloc./pivots & 0.00 & 0.00 & 0.00 & 0.00 \\
Spatial only & 75.21 & 70.71 & 75.92 & 73.95 \\
Temporal w/o pivots & 0.00 & 6.57 & 11.94 & 6.17 \\
Temporal only & 3.96 & 11.62 & 23.03 & 12.87 \\
\rowcolor{methodwash}
STEPQuant@4bit & \textbf{86.25} & \textbf{81.57} & \textbf{86.35} & \textbf{84.72} \\
\bottomrule
\end{tabular}
\end{minipage}

 \label{tab:ablation}
\end{table}

\noindent\textbf{Component analysis.}
(i) \textit{Spatial fitting.}
At 4 bits, our spatial fitting substantially outperforms
DSQ on Qwen (73.95\% vs. 7.64\%), with consistent
improvements at 6 bits.
Moreover, when using the same bit allocation and FP16 pivots, STEPQuant still outperforms Q-Mamba at both bit budgets, with a particularly large gap at 4 bits.
These results demonstrate the effectiveness of our key-row-aware dual-axis fitting.
(ii) \textit{Pivot protection.}
Protecting only 1.39\% of Qwen heads with FP16 pivots
improves the 4-bit three-task average by 6.70 points
and the 6-bit AIME accuracy by 14.16 points.
This demonstrates that protecting a small fraction
of high-risk units can substantially improve accuracy.
(iii) \textit{Combining both components.}
At 4 bits, STEPQuant achieves 84.72\% on Qwen,
outperforming both Spatial only (73.95\%) and
Temporal only (12.87\%).
At 6 bits, STEPQuant achieves 84.51\%, compared with
80.19\% and 75.63\% for the individual components,
closely matching FP32 (84.61\%).
These results demonstrate the complementary benefits
of our spatial and temporal components.
Similar trends are observed on Kimi under both bit
budgets.
\end{hbenv}

Additional controlled ablations of lifetime weighting, row-impact weighting, and dual-axis scaling are provided in Appendix~\ref{app:ablation}.

\begin{figure}[!t]
 \centering\includegraphics[width=0.95\linewidth]{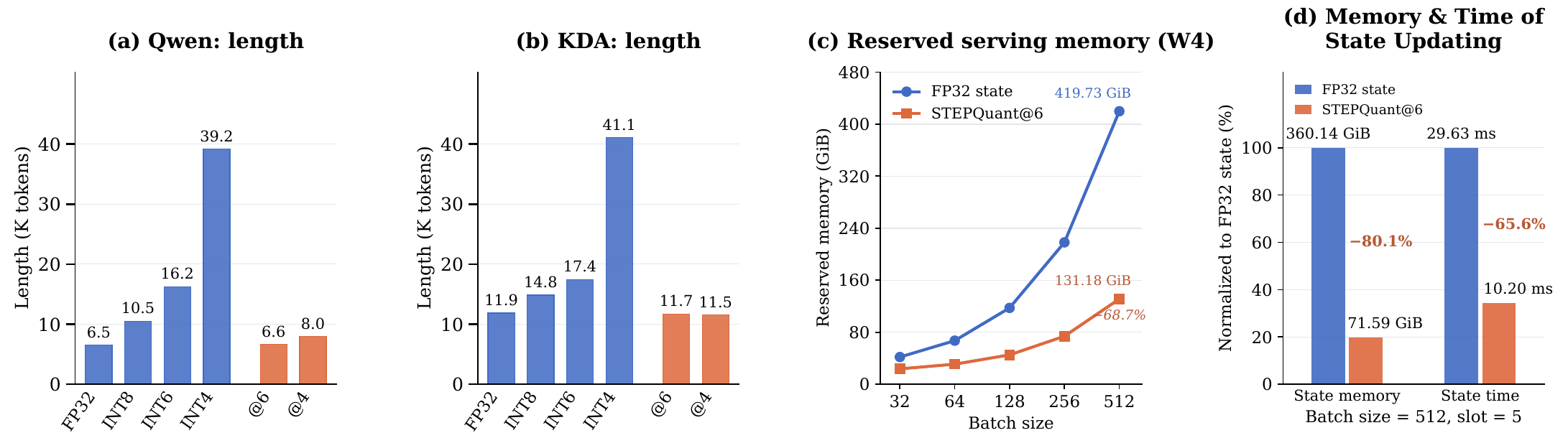}
\caption{\textbf{Generation length and serving efficiency of STEPQuant.}
 (a--b) Mean generated tokens across seven tasks, weighting tasks equally.
 Blue bars show FP32 and uniform INT8/6/4. Orange bars show
 STEPQuant@6bit/@4bit (marked @6/@4).
 Lengths include thinking, incorrect answers, and capped outputs.
 (c) Reserved serving-memory capacity of Qwen with W4 weights across different batch sizes.
 (d) Normalized recurrent-state memory and updating time of Qwen at batch size 512.
 }
 \label{fig:serving_throughput}
\end{figure}

\subsection{Generation length on seven long-generation tasks}
\label{sec:length}

\lin{
Prior works~\citep{liu2025quantization,lotfi2026quantized} have observed that quantized models may exhibit overthinking, producing excessively long outputs without improving reasoning accuracy.
We therefore examine generation length to further understand the behavior of STEPQuant on reasoning tasks.
As presented in Fig.~\ref{fig:serving_throughput}(a)(b), uniform quantization substantially increases output length.
For example, Kimi generates an average of 63.40K tokens on AIME and 64.22K on HMMT under 4-bit quantization, approaching the 65536-token limit while achieving near 0 accuracy on both tasks.
This suggests that the increased generation length does not translate into effective reasoning under aggressive uniform quantization.
In contrast, STEPQuant maintains output lengths close to those of the FP32-state model, consistent with its preserved accuracy on long reasoning tasks.
}

\subsection{Efficiency Evaluation}
\begin{hbenv}
\label{sec:serving_throughput}
We evaluate the serving efficiency of STEPQuant
using SGLang on A800 GPUs.
Fig.~\ref{fig:serving_throughput}(c) reports the reserved serving-memory capacity of Qwen with W4 weights across different batch sizes (8 $\times$ A800-80GB, TP8, $r=5$).
At a batch size of 512, STEPQuant@6bit reduces
reserved memory capacity
from 419.73 to 131.18 GiB, a 68.7\% reduction.
Beyond the overall memory savings, we further examine recurrent-state memory and state-update time.
Fig.~\ref{fig:serving_throughput}(d) shows that STEPQuant@6bit reduces recurrent-state memory by 80.1\% (5.03$\times$ compression) and state-update time by 65.6\% (2.91$\times$ faster).
Additional results, including the evaluation protocol, state memory and update time on Kimi, and full-model decode throughput on both models, are provided in Appendix~\ref{app:serving}.

\end{hbenv}
}

\section{Related Work}

Efficient LLM inference has been pursued through efficient decoding procedures~\citep{chen2026dflash,xu2026predict,qian2026adaflash,qian2026d3llm}, parameter pruning~\citep{zhang2024plug,lin2024mope,xing2025efficientllm,lin2026benchmarking}, low-bit quantization~\citep{frantar2022gptq,yang2026lrq,yang2026reshape,yang2026dapq,ma2023ompq,ma2023solving,ma2024outlier}, and architectural approaches~\citep{yang2023gla,yang2024parallelizing,lin2026efficient}. 
Our work combines the latter two directions by quantizing the recurrent states used in gated linear attention models. We therefore review these architectures and related PTQ methods below.

\subsection{Linear Attention Transformers}

\lin{
Standard self-attention incurs quadratic computation in sequence length and maintains a growing KV cache during autoregressive decoding~\citep{zhang2023h2o}. 
Linear attention can instead summarize past key-value interactions in a fixed-size recurrent state, enabling linear-time sequence processing and constant state storage with respect to sequence length. 
Subsequent works improve memory management and computational efficiency: RetNet~\citep{sun2023retent} introduces multiscale retention with parallel and recurrent formulations, while Gated Linear Attention~\citep{yang2023gla} uses data-dependent gates to control information retention. 
Related state-space models, including Mamba~\citep{gu2024mamba} and Mamba-2~\citep{dao2024transformers}, also combine recurrent inference with efficient training algorithms.
Gated DeltaNet (GDN)~\citep{yang2025gated} combines head-wise forgetting with delta-rule updates to selectively modify associations stored in its matrix-valued state. 
Kimi Delta Attention (KDA)~\citep{kimi2025linear} extends this mechanism with channel-wise forgetting and serves as the main component of Kimi Linear and K3~\citep{team2026kimi-k3}, which interleave KDA and softmax attention layers. 
Qwen families~\citep{qwen3.5,qwen3.6-27b,qwen3.6-35b-a3b,qwen2026qwen38} also adopt the hybrid GDN architecture to deliver high-throughput inference with minimal latency overhead.
These advances improve how recurrent states store and update information, but fixed state size does not eliminate storage costs: total state memory scales with batch size, layer count, and state dimensions. 
We study low-bit quantization of GDN recurrent states to reduce this inference memory footprint.
}

\subsection{Post-training Quantization}
\lin{
Post-training quantization (PTQ) reduces model memory by representing weights in low-bit formats~\citep{kim2023squeezellm,lin2024awq,shao2024omniquant,zhong2024erq,zhong2023s,zhong2025towards,zhang2026quantvla}, while jointly quantizing weights and activations can accelerate inference through low-bit kernels~\citep{ashkboos2024quarot,lin2024duquant,lin2026duquant++,lin2026quantization}. 
PTQ can also reduce the growing KV cache in long-context inference~\citep{liu2024intactkv,hooper2024kvquant,zandieh2026turboquant}.
For example, KIVI~\citep{liu2024kivi} quantizes keys and values along different axes. 
For state-space models, Quamba~\citep{chiang2025quamba} and MambaQuant~\citep{yue2025mambaquant} target weight and activation quantization, while Quamba2 \citep{chiang2025quamba2} additionally quantizes cached recurrent states to 8 bits. Q-Mamba~\citep{tianqi2025q} addresses state-cache quantization through dual-axis scaling and selectivity reconstruction.
Quantization of GDN recurrent states remains less explored. 
Concurrent work, DAMP~\citep{zhang2026damp}, uses decay-based persistence to select key channels for FP16 protection while quantizing the remaining channels to INT8, reporting preserved accuracy at 9.9 bits per state value in its evaluated settings.
Our work demonstrates that lower precision is feasible: \method\ quantizes recurrent states in KDA and Qwen3.8 models to 6 bits with negligible accuracy degradation on the evaluated benchmarks.
}
\section{Conclusion}
\lin{
We study low-bit quantization of recurrent states in Delta-rule models and show that quantization errors are shaped by both temporal persistence and spatial structure. Based on these observations, we propose STEPQuant, combining lifetime-aware bit allocation with Key-Row-Aware Dual-axis Fitting. Experiments across long- and short-generation tasks show that STEPQuant enables accurate low-bit state quantization and substantially reduces recurrent-state memory for concurrent serving.
}

\section*{AI Use Statement}
AI assistants were used only for language polishing, \LaTeX{} checking, and debugging assistance. All scientific ideas, methodological designs, experiments, analyses, and conclusions were developed, conducted, and verified by the authors.

\section*{Reproducibility Statement}
We describe every component needed to reproduce our experiments. 
We provide the recurrent-state update and quantization definitions in
Section~\ref{sec:preliminaries}, the precision-allocation objective in
Section~\ref{sec:lmsq}, and the spatial fitting procedure in
Section~\ref{sec:fitting}. Appendix~\ref{app:proofs} gives the proof of
conditional error propagation and defines the calibration statistics.
Appendices~\ref{app:calibration} and~\ref{app:algorithm} describe the
calibration data, candidate precisions, FP16 pivots, allocation procedures,
and quantized decode workflow.
Section~\ref{sec:protocol} specifies the models, benchmarks, and generation
settings, while Appendix~\ref{app:serving} documents the SGLang integration,
hardware configurations, decode timing protocol, and memory accounting.
Additional per-task generation lengths and component ablations are
provided in Appendix~\ref{app:provenance}.

\section*{Future Work.}
Recent work explores several directions for efficient linear recurrent models, including state quantization~\citep{pan2026leapquant,nazari2026lores}, low-precision prefill~\citep{tumma2026linquant}, and reduced state-access overhead~\citep{liou2026replayssm,lin2026deltalog}.
Sparse computation~\citep{zhuang2026sparsegdn}, state-dimension reduction~\citep{nazari2026state}, and efficient speculative verification~\citep{ghantasala2026treewy} offer further opportunities.
An important next step is to study how these techniques interact with state quantization, particularly their effects on error accumulation and end-to-end inference efficiency.
We hope that our error analysis and open-source implementation help others investigate these questions and develop more efficient inference methods for GDN and related recurrent models.

\bibliographystyle{iclr2027_conference}
\bibliography{references}

@inproceedings{yang2025gated,
  title={Gated delta networks: Improving mamba2 with delta rule},
  author={Yang, Songlin and Kautz, Jan and Hatamizadeh, Ali},
  booktitle={International Conference on Learning Representations},
  volume={2025},
  pages={29687--29707},
  year={2025}
}

@article{kimi2025linear,
  title={Kimi linear: An expressive, efficient attention architecture},
  author={Team, Kimi and Zhang, Yu and Lin, Zongyu and Yao, Xingcheng and Hu, Jiaxi and Meng, Fanqing and Liu, Chengyin and Men, Xin and Yang, Songlin and Li, Zhiyuan and others},
  journal={arXiv preprint arXiv:2510.26692},
  year={2025}
}

@inproceedings{lin2024mope,
  title={Mope-clip: Structured pruning for efficient vision-language models with module-wise pruning error metric},
  author={Lin, Haokun and Bai, Haoli and Liu, Zhili and Hou, Lu and Sun, Muyi and Song, Linqi and Wei, Ying and Sun, Zhenan},
  booktitle={Proceedings of the IEEE/CVF Conference on Computer Vision and Pattern Recognition},
  pages={27370--27380},
  year={2024}
}

@article{dao2024transformers,
  title={Transformers are SSMs: Generalized models and efficient algorithms through structured state space duality},
  author={Dao, Tri and Gu, Albert},
  journal={arXiv preprint arXiv:2405.21060},
  year={2024}
}

@article{ashkboos2024quarot,
  title={Quarot: Outlier-free 4-bit inference in rotated llms},
  author={Ashkboos, Saleh and Mohtashami, Amirkeivan and Croci, Maximilian L and Li, Bo and Cameron, Pashmina and Jaggi, Martin and Alistarh, Dan and Hoefler, Torsten and Hensman, James},
  journal={Advances in Neural Information Processing Systems},
  volume={37},
  pages={100213--100240},
  year={2024}
}

@inproceedings{chiang2025quamba,
  title={Quamba: A post-training quantization recipe for selective state space models},
  author={Chiang, Hung-Yueh and Chang, Chi-Chih and Frumkin, Natalia and Wu, Kai-Chiang and Marculescu, Diana},
  booktitle={International Conference on Learning Representations},
  volume={2025},
  pages={101328--101354},
  year={2025}
}

@article{liu2024kivi,
  title={Kivi: A tuning-free asymmetric 2bit quantization for kv cache},
  author={Liu, Zirui and Yuan, Jiayi and Jin, Hongye and Zhong, Shaochen and Xu, Zhaozhuo and Braverman, Vladimir and Chen, Beidi and Hu, Xia},
  journal={arXiv preprint arXiv:2402.02750},
  year={2024}
}

@article{zhang2026damp,
  title={DAMP: Decay-Aware Mixed-Precision Recurrent-State Quantization},
  author={Zhang, Tao and Tan, Jianchao and Sun, Pingwei and Yu, Yanqi and Jiang, Zixu and Xie, Yuchen and Cai, Xunliang and Zeng, Ziqian},
  journal={arXiv preprint arXiv:2608.27513},
  year={2026}
}

@inproceedings{jain2025livecodebench,
  title={Livecodebench: Holistic and contamination free evaluation of large language models for code},
  author={Jain, Naman and Han, King and Gu, Alex and Li, Wen-Ding and Yan, Fanjia and Zhang, Tianjun and Wang, Sida and Solar-Lezama, Armando and Sen, Koushik and Stoica, Ion},
  booktitle={International Conference on Learning Representations},
  volume={2025},
  pages={58791--58831},
  year={2025}
}

@article{rein2023gpqa,
  title={Gpqa: A graduate-level google-proof q\&a benchmark},
  author={Rein, David and Hou, Betty Li and Stickland, Asa Cooper and Petty, Jackson and Pang, Richard Yuanzhe and Dirani, Julien and Michael, Julian and Bowman, Samuel R},
  journal={arXiv preprint arXiv:2311.12022},
  year={2023}
}

@inproceedings{zellers2019hellaswag,
  title={Hellaswag: Can a machine really finish your sentence?},
  author={Zellers, Rowan and Holtzman, Ari and Bisk, Yonatan and Farhadi, Ali and Choi, Yejin},
  booktitle={Proceedings of the 57th annual meeting of the association for computational linguistics},
  pages={4791--4800},
  year={2019}
}

@article{sakaguchi2021winogrande,
  title={Winogrande: An adversarial winograd schema challenge at scale},
  author={Sakaguchi, Keisuke and Bras, Ronan Le and Bhagavatula, Chandra and Choi, Yejin},
  journal={Communications of the ACM},
  volume={64},
  number={9},
  pages={99--106},
  year={2021},
  publisher={ACM New York, NY, USA}
}

@inproceedings{paperno2016lambada,
  title={The LAMBADA dataset: Word prediction requiring a broad discourse context},
  author={Paperno, Denis and Kruszewski, Germ{\'a}n and Lazaridou, Angeliki and Pham, Ngoc-Quan and Bernardi, Raffaella and Pezzelle, Sandro and Baroni, Marco and Boleda, Gemma and Fern{\'a}ndez, Raquel},
  booktitle={Proceedings of the 54th annual meeting of the association for computational linguistics (volume 1: Long papers)},
  pages={1525--1534},
  year={2016}
}

@article{liu2023your,
  title={Is your code generated by chatgpt really correct? rigorous evaluation of large language models for code generation},
  author={Liu, Jiawei and Xia, Chunqiu Steven and Wang, Yuyao and Zhang, Lingming},
  journal={Advances in neural information processing systems},
  volume={36},
  pages={21558--21572},
  year={2023}
}

@inproceedings{
pyatkin2026generalizing,
title={Generalizing Verifiable Instruction Following},
author={Valentina Pyatkin and Saumya Malik and Victoria Graf and Hamish Ivison and Shengyi Huang and Pradeep Dasigi and Nathan Lambert and Hannaneh Hajishirzi},
booktitle={The Thirty-ninth Annual Conference on Neural Information Processing Systems Datasets and Benchmarks Track},
year={2026},
url={https://openreview.net/forum?id=yfYgwjj5F8}
}

@article{hendrycks2020measuring,
  title={Measuring massive multitask language understanding},
  author={Hendrycks, Dan and Burns, Collin and Basart, Steven and Zou, Andy and Mazeika, Mantas and Song, Dawn and Steinhardt, Jacob},
  journal={arXiv preprint arXiv:2009.03300},
  year={2020}
}

@article{chiang2025quamba2,
  title={Quamba2: A robust and scalable post-training quantization framework for selective state space models},
  author={Chiang, Hung-Yueh and Chang, Chi-Chih and Frumkin, Natalia and Wu, Kai-Chiang and Abdelfattah, Mohamed S and Marculescu, Diana},
  journal={arXiv preprint arXiv:2503.22879},
  year={2025}
}

@inproceedings{tianqi2025q,
  title={Q-Mamba: Towards more efficient Mamba models via post-training quantization},
  author={Tianqi, Chen and Chen, Yuanteng and Wang, Peisong and Xu, Weixiang and Zhu, Zeyu and Cheng, Jian},
  booktitle={Findings of the Association for Computational Linguistics: ACL 2025},
  pages={10594--10610},
  year={2025}
}

@article{zheng2024sglang,
  title={Sglang: Efficient execution of structured language model programs},
  author={Zheng, Lianmin and Yin, Liangsheng and Xie, Zhiqiang and Sun, Chuyue and Huang, Jeff and Yu, Cody H and Cao, Shiyi and Kozyrakis, Christos and Stoica, Ion and Gonzalez, Joseph E and others},
  journal={Advances in neural information processing systems},
  volume={37},
  pages={62557--62583},
  year={2024}
  
}

@article{raffel2020exploring,
  title={Exploring the limits of transfer learning with a unified text-to-text transformer},
  author={Raffel, Colin and Shazeer, Noam and Roberts, Adam and Lee, Katherine and Narang, Sharan and Matena, Michael and Zhou, Yanqi and Li, Wei and Liu, Peter J},
  journal={Journal of machine learning research},
  volume={21},
  number={140},
  pages={1--67},
  year={2020}
}

@article{merity2016pointer,
  title={Pointer sentinel mixture models},
  author={Merity, Stephen and Xiong, Caiming and Bradbury, James and Socher, Richard},
  journal={arXiv preprint arXiv:1609.07843},
  year={2016}
}

@article{dekoninck2026beyond,
  title={Beyond benchmarks: Matharena as an evaluation platform for mathematics with llms},
  author={Dekoninck, Jasper and Jovanovi{\'c}, Nikola and Gehrunger, Tim and R{\"o}gnvaldsson, K{\'a}ri and Petrov, Ivo and Sun, Chenhao and Vechev, Martin},
  journal={arXiv preprint arXiv:2605.00674},
  year={2026}
}

@inproceedings{lightman2024let,
  title={Let's verify step by step},
  author={Lightman, Hunter and Kosaraju, Vineet and Burda, Yuri and Edwards, Harrison and Baker, Bowen and Lee, Teddy and Leike, Jan and Schulman, John and Sutskever, Ilya and Cobbe, Karl},
  booktitle={International Conference on Learning Representations},
  volume={2024},
  pages={39578--39601},
  year={2024}
}

@article{clark2018think,
  title={Think you have solved question answering? try arc, the ai2 reasoning challenge},
  author={Clark, Peter and Cowhey, Isaac and Etzioni, Oren and Khot, Tushar and Sabharwal, Ashish and Schoenick, Carissa and Tafjord, Oyvind},
  journal={arXiv preprint arXiv:1803.05457},
  year={2018}
}

@inproceedings{mihaylov2018can,
  title={Can a suit of armor conduct electricity? a new dataset for open book question answering},
  author={Mihaylov, Todor and Clark, Peter and Khot, Tushar and Sabharwal, Ashish},
  booktitle={Proceedings of the 2018 conference on empirical methods in natural language processing},
  pages={2381--2391},
  year={2018}
}

@article{yang2023gla,
  title={Gated linear attention transformers with hardware-efficient training},
  author={Yang, Songlin and Wang, Bailin and Shen, Yikang and Panda, Rameswar and Kim, Yoon},
  journal={arXiv preprint arXiv:2312.06635},
  year={2023}
}

@article{team2026kimi-k3,
  title={Kimi k3: Open frontier intelligence},
  author={Team, Kimi and Bai, Tongtong and Bai, Yifan and Bao, Yiping and Cai, Jianfeng and Cai, Xinyuan and Cao, Peizhou and Cao, Yuxuan and Chai, Ziwei and Charles, Y and others},
  journal={arXiv preprint arXiv:2607.24653},
  year={2026}
}

@inproceedings{yue2025mambaquant,
  title={Mambaquant: Quantizing the mamba family with variance aligned rotation methods},
  author={Xu, Zukang and Yue, Yuxuan and Hu, Xing and Yuan, Zhihang and Jiang, Zixu and Chen, Zhixuan and Yu, Jiangyong and Xu, Chen and Zhou, Sifan and Yang, Dawei},
  booktitle={International Conference on Learning Representations},
  volume={2025},
  pages={33231--33250},
  year={2025}
}

@inproceedings{zhang2024plug,
  title={Plug-and-play: An efficient post-training pruning method for large language models},
  author={Zhang, Yingtao and Bai, Haoli and Lin, Haokun and Zhao, Jialin and Hou, Lu and Cannistraci, Carlo Vittorio},
  booktitle={International Conference on Learning Representations},
  volume={2024},
  pages={50490--50508},
  year={2024}
}

@article{xing2025efficientllm,
  title={Efficientllm: Scalable pruning-aware pretraining for architecture-agnostic edge language models},
  author={Xing, Xingrun and Liu, Zheng and Xiao, Shitao and Gao, Boyan and Liang, Yiming and Zhang, Wanpeng and Lin, Haokun and Li, Guoqi and Zhang, Jiajun},
  journal={arXiv preprint arXiv:2502.06663},
  year={2025}
}

@article{hooper2024kvquant,
  title={Kvquant: Towards 10 million context length llm inference with kv cache quantization},
  author={Hooper, Coleman and Kim, Sehoon and Mohammadzadeh, Hiva and Mahoney, Michael W and Shao, Yakun S and Keutzer, Kurt and Gholami, Amir},
  journal={Advances in Neural Information Processing Systems},
  volume={37},
  pages={1270--1303},
  year={2024}
}

@article{lin2024awq,
  title={Awq: Activation-aware weight quantization for on-device llm compression and acceleration},
  author={Lin, Ji and Tang, Jiaming and Tang, Haotian and Yang, Shang and Chen, Wei-Ming and Wang, Wei-Chen and Xiao, Guangxuan and Dang, Xingyu and Gan, Chuang and Han, Song},
  journal={Proceedings of machine learning and systems},
  volume={6},
  pages={87--100},
  year={2024}
}

@inproceedings{shao2024omniquant,
  title={Omniquant: Omnidirectionally calibrated quantization for large language models},
  author={Shao, Wenqi and Chen, Mengzhao and Zhang, Zhaoyang and Xu, Peng and Zhao, Lirui and Li, Zhiqian and Zhang, Kaipeng and Peng, Gao and Qiao, Yu and Luo, Ping},
  booktitle={International Conference on Learning Representations},
  volume={2024},
  pages={45472--45496},
  year={2024}
}

@article{lin2024duquant,
  title={Duquant: Distributing outliers via dual transformation makes stronger quantized llms},
  author={Lin, Haokun and Xu, Haobo and Wu, Yichen and Cui, Jingzhi and Zhang, Yingtao and Mou, Linzhan and Song, Linqi and Sun, Zhenan and Wei, Ying},
  journal={Advances in Neural Information Processing Systems},
  volume={37},
  pages={87766--87800},
  year={2024}
}

@inproceedings{zandieh2026turboquant,
  title={Turboquant: Online vector quantization with near-optimal distortion rate},
  author={Zandieh, Amir and Daliri, Majid and Hadian, Majid and Mirrokni, Vahab},
  booktitle={International Conference on Learning Representations},
  volume={2026},
  pages={56418--56439},
  year={2026}
}

@article{dettmers2022gpt3,
  title={Gpt3. int8 (): 8-bit matrix multiplication for transformers at scale},
  author={Dettmers, Tim and Lewis, Mike and Belkada, Younes and Zettlemoyer, Luke},
  journal={Advances in neural information processing systems},
  volume={35},
  pages={30318--30332},
  year={2022}
}

@inproceedings{xiao2023smoothquant,
  title={Smoothquant: Accurate and efficient post-training quantization for large language models},
  author={Xiao, Guangxuan and Lin, Ji and Seznec, Mickael and Wu, Hao and Demouth, Julien and Han, Song},
  booktitle={International conference on machine learning},
  pages={38087--38099},
  year={2023},
  organization={PMLR}
}

@article{box1964analysis,
  title={An analysis of transformations},
  author={Box, George EP and Cox, David R},
  journal={Journal of the Royal Statistical Society Series B: Statistical Methodology},
  volume={26},
  number={2},
  pages={211--243},
  year={1964},
  publisher={Oxford University Press}
}

@inproceedings{bai2024longbench,
  title={Longbench: A bilingual, multitask benchmark for long context understanding},
  author={Bai, Yushi and Lv, Xin and Zhang, Jiajie and Lyu, Hongchang and Tang, Jiankai and Huang, Zhidian and Du, Zhengxiao and Liu, Xiao and Zeng, Aohan and Hou, Lei and others},
  booktitle={Proceedings of the 62nd annual meeting of the association for computational linguistics (volume 1: Long papers)},
  pages={3119--3137},
  year={2024}
}

@article{yang2026reshape,
  title={Reshape and rotate: Adaptive weight reshaping and fine-grained rotation for ultra-low-bit diffusion transformers quantization},
  author={Yang, Lianwei and Lin, Haokun and Wu, Yichen and Shan, Caifeng and Sun, Zhenan and Gu, Qingyi},
  journal={Neurocomputing},
  pages={133830},
  year={2026},
  publisher={Elsevier}
}

@article{lin2026benchmarking,
  title={Benchmarking Trustworthiness of SLMs: Pre-trained vs. Compressed},
  author={Lin, Haokun and Zhu, Kaijie and Xu, Haobo and Wu, Yichen and Lu, Zhichao and Zhang, Qingfu and Sun, Zhenan},
  journal={arXiv preprint arXiv:2608.11981},
  year={2026}
}

@misc{qwen3.6-27b,
    title  = {{Qwen3.6-27B}: Flagship-Level Coding in a {27B} Dense Model},
    author = {{Qwen Team}},
    year   = {2026},
    month  = {April},
    url    = {https://qwen.ai/blog?id=qwen3.6-27b}
}

@misc{qwen3.6-35b-a3b,
    title  = {{Qwen3.6-35B-A3B}: Agentic Coding Power, Now Open to All},
    author = {{Qwen Team}},
    year   = {2026},
    month  = {April},
    url    = {https://qwen.ai/blog?id=qwen3.6-35b-a3b}
}

@misc{qwen3.5,
    title  = {{Qwen3.5}: Towards Native Multimodal Agents},
    author = {{Qwen Team}},
    year   = {2026},
    month  = {February},
    url    = {https://qwen.ai/blog?id=qwen3.5}
}

@article{sun2023retent,
  title={Retentive network: A successor to transformer for large language models},
  author={Sun, Yutao and Dong, Li and Huang, Shaohan and Ma, Shuming and Xia, Yuqing and Xue, Jilong and Wang, Jianyong and Wei, Furu},
  journal={arXiv preprint arXiv:2307.08621},
  year={2023}
}

@article{frantar2022gptq,
  title={Gptq: Accurate post-training quantization for generative pre-trained transformers},
  author={Frantar, Elias and Ashkboos, Saleh and Hoefler, Torsten and Alistarh, Dan},
  journal={arXiv preprint arXiv:2210.17323},
  year={2022}
}

@article{lin2026duquant++,
  title={DuQuant++: Fine-grained Rotation Enhances Microscaling FP4 Quantization},
  author={Lin, Haokun and Jia, Xinle and Xu, Haobo and Yao, Bingchen and Guo, Xianglong and Wu, Yichen and Lu, Zhichao and Wei, Ying and Zhang, Qingfu and Sun, Zhenan},
  journal={arXiv preprint arXiv:2604.17789},
  year={2026}
}

@inproceedings{ma2023ompq,
  title={Ompq: Orthogonal mixed precision quantization},
  author={Ma, Yuexiao and Jin, Taisong and Zheng, Xiawu and Wang, Yan and Li, Huixia and Wu, Yongjian and Jiang, Guannan and Zhang, Wei and Ji, Rongrong},
  booktitle={Proceedings of the AAAI conference on artificial intelligence},
  volume={37},
  number={7},
  pages={9029--9037},
  year={2023}
}

@inproceedings{ma2024outlier,
  title={Outlier-aware slicing for post-training quantization in vision transformer},
  author={Ma, Yuexiao and Li, Huixia and Zheng, Xiawu and Ling, Feng and Xiao, Xuefeng and Wang, Rui and Wen, Shilei and Chao, Fei and Ji, Rongrong},
  booktitle={Forty-first International Conference on Machine Learning},
  year={2024}
}

@INPROCEEDINGS{ma2023solving,
  author={Ma, Yuexiao and Li, Huixia and Zheng, Xiawu and Xiao, Xuefeng and Wang, Rui and Wen, Shilei and Pan, Xin and Chao, Fei and Ji, Rongrong},
  booktitle={2023 IEEE/CVF Conference on Computer Vision and Pattern Recognition (CVPR)}, 
  title={Solving Oscillation Problem in Post-Training Quantization Through a Theoretical Perspective}, 
  year={2023},
  volume={},
  number={},
  pages={7950-7959},
  doi={10.1109/CVPR52729.2023.00768}}

@article{lin2026efficient,
  title={Efficient diffusion language models: A comprehensive survey},
  author={Lin, Haokun and Jia, Xinle and Liu, Shaozhen and Xia, Shujun and Huang, Weitao and Xu, Haobo and Li, Junyang and Xiao, Yicheng and Xing, Xingrun and Guo, Ziyu and others},
  year={2026},
  publisher={Authorea}
}

@article{lin2026quantization,
  title={Quantization meets dllms: A systematic study of post-training quantization for diffusion llms},
  author={Lin, Haokun and Xu, Haobo and Wu, Yichen and Guo, Ziyu and Zhang, Renrui and Lu, Zhichao and Wei, Ying and Zhang, Qingfu and Sun, Zhenan},
  journal={Machine Intelligence Research},
  pages={1--17},
  year={2026},
  publisher={Springer}
}

@inproceedings{yang2026dapq,
  title={DapQ-DiT: Distribution-Aware Post-Training Quantization for Efficient Generative Tasks in Diffusion Transformers},
  author={Yang, Lianwei and Lin, Haokun and Wu, Yichen and Sun, Zhenan and Gu, Qingyi},
  booktitle={Proceedings of the 2026 International Conference on Multimedia Retrieval},
  pages={2371--2380},
  year={2026}
}

@article{yang2026lrq,
  title={Lrq-dit: Log-rotation post-training quantization of diffusion transformers for image and video generation},
  author={Yang, Lianwei and Lin, Haokun and Zhao, Tianchen and Wu, Yichen and Zhu, Hongyu and Xie, Ruiqi and Sun, Zhenan and Wang, Yu and Gu, Qingyi},
  journal={IEEE Transactions on Circuits and Systems for Video Technology},
  year={2026},
  publisher={IEEE}
}

@article{zhang2026quantvla,
  title={Quantvla: Scale-calibrated post-training quantization for vision-language-action models},
  author={Zhang, Jingxuan and Hsieh, Yunta and Wan, Zhongwei and Lin, Haokun and Wang, Xin and Wang, Ziqi and Lei, Yingtie and Zhang, Mi},
  journal={arXiv preprint arXiv:2602.20309},
  year={2026}
}

@inproceedings{gu2024mamba,
 author = {Gu, Albert and Dao, Tri},
 booktitle = {First Conference on Language Modeling},
 title = {{Mamba}: Linear-Time Sequence Modeling with Selective State Spaces},
 url = {https://openreview.net/forum?id=tEYskw1VY2},
 year = {2024}
}

@misc{qwen2026qwen38,
    title  = {{Qwen3.8-Max}: A New Bar for Coding and Cowork},
    author = {{Qwen Team}},
    year   = {2026},
    month  = {August},
    url    = {https://qwen.ai/blog?id=qwen3.8}
}

@article{dettmers2023spqr,
  title={Spqr: A sparse-quantized representation for near-lossless llm weight compression},
  author={Dettmers, Tim and Svirschevski, Ruslan and Egiazarian, Vage and Kuznedelev, Denis and Frantar, Elias and Ashkboos, Saleh and Borzunov, Alexander and Hoefler, Torsten and Alistarh, Dan},
  journal={arXiv preprint arXiv:2306.03078},
  year={2023}
}

@article{zhang2023h2o,
  title={H2o: Heavy-hitter oracle for efficient generative inference of large language models},
  author={Zhang, Zhenyu and Sheng, Ying and Zhou, Tianyi and Chen, Tianlong and Zheng, Lianmin and Cai, Ruisi and Song, Zhao and Tian, Yuandong and R{\'e}, Christopher and Barrett, Clark and others},
  journal={Advances in neural information processing systems},
  volume={36},
  pages={34661--34710},
  year={2023}
}

@article{lotfi2026quantized,
  title={Quantized reasoning models think they need to think longer, but they do not},
  author={Lotfi, Sanae and Kirichenko, Polina and Li, Steven and Liu, Zechun},
  journal={arXiv preprint arXiv:2606.00206},
  year={2026}
}

@article{liu2024intactkv,
  title={Intactkv: Improving large language model quantization by keeping pivot tokens intact},
  author={Liu, Ruikang and Bai, Haoli and Lin, Haokun and Li, Yuening and Gao, Han and Xu, Zhengzhuo and Hou, Lu and Yao, Jun and Yuan, Chun},
  journal={arXiv preprint arXiv:2403.01241},
  year={2024}
}

@article{liu2025quantization,
  title={Quantization hurts reasoning? an empirical study on quantized reasoning models},
  author={Liu, Ruikang and Sun, Yuxuan and Zhang, Manyi and Bai, Haoli and Yu, Xianzhi and Yu, Tiezheng and Yuan, Chun and Hou, Lu},
  journal={arXiv preprint arXiv:2504.04823},
  year={2025}
}

@article{xu2026predict,
  title={Predict, Don't Iterate: Efficient Adaptive-Length Infilling for Diffusion Language Models},
  author={Xu, Haobo and Chen, Sirui and Bei, Yuanchen and Chen, Lingjie and Yan, Yuchen and Fu, Dongqi and He, Jingrui and Tong, Hanghang},
  journal={arXiv preprint arXiv:2609.02108},
  year={2026}
}

@article{chen2026dflash,
  title={Dflash: Block diffusion for flash speculative decoding},
  author={Chen, Jian and Liang, Yesheng and Liu, Zhijian},
  journal={arXiv preprint arXiv:2602.06036},
  year={2026}
}

@article{qian2026adaflash,
  title={AdaFlash: Adaptive Speculative Decoding via On-Policy Distilled Diffusion Drafters},
  author={Qian, Yu-Yang and Wu, Hao-Cong and Chen, Chen and Sun, Jiacheng and Dong, Zhenhua and Zhao, Peng and Zhou, Zhi-Hua},
  journal={arXiv preprint arXiv:2607.19223},
  year={2026}
}

@article{qian2026d3llm,
  title={d3llm: Ultra-fast diffusion llm using pseudo-trajectory distillation},
  author={Qian, Yu-Yang and Su, Junda and Hu, Lanxiang and Zhang, Peiyuan and Deng, Zhijie and Zhao, Peng and Zhang, Hao},
  journal={arXiv preprint arXiv:2601.07568},
  year={2026}
}

@article{kim2023squeezellm,
  title={Squeezellm: Dense-and-sparse quantization},
  author={Kim, Sehoon and Hooper, Coleman and Gholami, Amir and Dong, Zhen and Li, Xiuyu and Shen, Sheng and Mahoney, Michael W and Keutzer, Kurt},
  journal={arXiv preprint arXiv:2306.07629},
  year={2023}
}

@article{yang2024parallelizing,
  title={Parallelizing linear transformers with the delta rule over sequence length},
  author={Yang, Songlin and Wang, Bailin and Zhang, Yu and Shen, Yikang and Kim, Yoon},
  journal={Advances in neural information processing systems},
  volume={37},
  pages={115491--115522},
  year={2024}
}

@article{zhong2023s,
  title={I\&s-vit: An inclusive \& stable method for pushing the limit of post-training vits quantization},
  author={Zhong, Yunshan and Hu, Jiawei and Lin, Mingbao and Chen, Mengzhao and Ji, Rongrong},
  journal={IEEE Transactions on Pattern Analysis \& Machine Intelligence (TPAMI)},
  doi={10.1109/TPAMI.2025.3610466},
  year={2025}
}

@article{zhong2025towards,
  title={Towards accurate post-training quantization of vision transformers via error reduction},
  author={Zhong, Yunshan and Huang, You and Hu, Jiawei and Zhang, Yuxin and Ji, Rongrong},
  journal={IEEE Transactions on Pattern Analysis and Machine Intelligence},
  year={2025},
  publisher={IEEE}
}

@inproceedings{zhong2024erq,
  title={Erq: Error reduction for post-training quantization of vision transformers},
  author={Zhong, Yunshan and Hu, Jiawei and Huang, You and Zhang, Yuxin and Ji, Rongrong},
  booktitle={Forty-first International Conference on Machine Learning},
  year={2024}
}

@article{hsieh2024ruler,
  title={RULER: What's the real context size of your long-context language models?},
  author={Hsieh, Cheng-Ping and Sun, Simeng and Kriman, Samuel and Acharya, Shantanu and Rekesh, Dima and Jia, Fei and Zhang, Yang and Ginsburg, Boris},
  journal={arXiv preprint arXiv:2404.06654},
  year={2024}
}

@article{pan2026leapquant,
  title   = {{LeapQuant}: Efficient Linear Attention with Accurate Recurrent State Quantization},
  author  = {Pan, Yi and Xi, Haocheng and Zhu, Kan and Li, Xingyang and Wu, Yibo and Mishra, Mayank and Zhang, Hongtao and Zheng, William X. and Kasikci, Baris and Han, Song and Keutzer, Kurt and Iyer, Rishabh and Stoica, Ion},
  journal = {arXiv preprint arXiv:2609.38166},
  year    = {2026}
}

@article{nazari2026lores,
  title   = {Quantizing Recurrent Language Models Is Easier Than You Think},
  author  = {Nazari, Philipp and Tumma, Neehal and Rus, Daniela and Rusch, T. Konstantin},
  journal = {arXiv preprint arXiv:2611.01864},
  year    = {2026}
}

@article{tumma2026linquant,
  title   = {{LinQuant}: Functional Invariances for Low-Precision Linear Recurrences},
  author  = {Tumma, Neehal and Nazari, Philipp and Cherukuri, Rohan and Rus, Daniela and Rusch, T. Konstantin},
  journal = {arXiv preprint arXiv:2610.39418},
  year    = {2026}
}

@article{liou2026replayssm,
  title   = {{ReplaySSM}: Cache {SSM} Inputs, Not State},
  author  = {Liou, Ze-Wei and Nover, Harris and Naik, Amey and Skliar, Andrii and Shovkun, Igor and Shahaf, Ido and Dao, Tri},
  journal = {arXiv preprint arXiv:2610.13778},
  year    = {2026}
}

@article{lin2026deltalog,
  title   = {{DeltaLog}: Deferred Materialization of Recurrent States for Linear Attention Decoding},
  author  = {Lin, Junqing and Sun, Jingwei and Sun, Guangzhong},
  journal = {arXiv preprint arXiv:2608.15533},
  year    = {2026}
}

@article{zhuang2026sparsegdn,
  title   = {{SparseGDN}: Chunk Sparse in Linear Attention},
  author  = {Zhuang, Jiedong and Zhao, Peng and Tang, Jialin and Hu, Haoji},
  journal = {arXiv preprint arXiv:2610.41420},
  year    = {2026}
}

@article{nazari2026state,
  title   = {On State Reduction in Linear Attention},
  author  = {Nazari, Philipp and Rusch, T. Konstantin},
  journal = {arXiv preprint arXiv:2602.04852},
  year    = {2026}
}

@article{ghantasala2026treewy,
  title   = {{TreeWY}: Speculative Verification for {Gated DeltaNet} Hybrids},
  author  = {Ghantasala, Sneha Murthy},
  journal = {arXiv preprint arXiv:2608.20961},
  year    = {2026}
}
\clearpage
\appendix
\renewcommand\thefigure{\Alph{section}\arabic{figure}}
\renewcommand\thetable{\Alph{section}\arabic{table}}
\setcounter{figure}{0}
\setcounter{table}{0}

\section{Recurrent-State Error Propagation and Calibration Statistics}
\label{app:proofs}
This appendix proves the conditional error propagation in
Section~\ref{sec:analysis} and relates cumulative squared output error
to the row-impact scores used by Key-Row-Aware Dual-axis Fitting
in Section~\ref{sec:fitting}. It also defines the calibration statistics
used by STEPQuant's temporal and spatial components.

\subsection{Proof of Proposition~\ref{prop:error}}
Subtracting the reference recurrence from the quantized update gives
\begin{align*}
 E_t&=\mathcal Q_t(A_t\widehat S_{t-1}+B_t)-(A_tS_{t-1}+B_t)\\
 &=A_t(\widehat S_{t-1}-S_{t-1})+\varepsilon_t.
\end{align*}
At step $t$, the output is read from $X_t$ before $\mathcal Q_t(X_t)$
is stored, so
$\widehat y_t-y_t=(A_tE_{t-1})^\top q_t$.
Let $P_t=I-\beta_tk_tk_t^\top$. Under the proposition's conditions,
$0\preceq P_t\preceq I$, so $\norm{P_t}_2\leq1$. Therefore
\[
 \norm{A_t}_2=\norm{P_tD_t}_2
 \leq\norm{P_t}_2\norm{D_t}_2\leq\norm{D_t}_2\leq1.
\]

\subsection{Cumulative squared output error}
\label{a.2}
Condition on the same keys, values, queries, and gates in both paths,
as in Proposition~\ref{prop:error}.
Let $\Phi_{s,t}=A_s\cdots A_{t+1}$ transport a state perturbation from
step $t$ to $s$. Over a horizon of $H$ future reads, an isolated quantization
error $Z$ produces cumulative squared output error
\begin{equation}
 \sum_{s=t+1}^{t+H}\|Z^\top\Phi_{s,t}^\top q_s\|_2^2
 =\tr(Z^\top W_{t,H}Z),\quad
 W_{t,H}=\sum_{s=t+1}^{t+H}\Phi_{s,t}^\top q_sq_s^\top\Phi_{s,t}.
 \label{eq:gramian}
\end{equation}
For a single perturbation $Z$ at time $t$ and no subsequent quantization errors,
the state perturbation after transition $s$ is $\Phi_{s,t}Z$.
The output perturbation is $Z^\top\Phi_{s,t}^\top q_s$. Using
$\norm{Z^\top u}_2^2=\tr(Z^\top uu^\top Z)$ and summing proves
Equation~\ref{eq:gramian}. Here $W_{t,H}$ is the readout-error Gram matrix and $\tr$ denotes
the matrix trace. This matrix is positive semidefinite.

For a single future read, $H=1$, this matrix reduces to
\[
 W_{t,1}=g_{t+1}g_{t+1}^\top,\qquad
 g_{t+1}=A_{t+1}^\top q_{t+1}.
\]
A unit-Frobenius-norm perturbation confined to key row $i$ therefore
produces squared output error at the next step $g_{t+1,i}^2$.
Both Qwen and KDA calibrate
$\omega_i=\E_{\rm cal}[g_{t,i}^2]$, the mean diagonal of this one-step
readout-error matrix. STEPQuant's spatial component normalizes
$\omega_i$ into the row-impact factors $w_i$. These determine the row
scales $r_i=m_i^{1/2}w_i^{-1/2}$ and the squared weights $w_i^2$ in the
column-fitting objective (Section~\ref{sec:fitting} and
Appendix~\ref{app:method_details}).

With quantization errors $\varepsilon_r$ introduced at successive steps
and $E_0=0$, the accumulated error is
\[
 E_t=\sum_{r=1}^{t}\Phi_{t,r}\varepsilon_r,
 \qquad\Phi_{t,t}=I.
\]
The squared norm includes cross-step terms between quantization errors introduced at different decoding steps. Separately, the quadratic form in Eq.~\eqref{eq:gramian} also contains cross-row interactions between different key rows of a single perturbation \(Z\), which we examine below.

\paragraph{Cross-row interactions.}
The quadratic form in Eq.~\eqref{eq:gramian} can be decomposed into
diagonal and cross-row terms:
\begin{equation}
\begin{aligned}
\operatorname{tr}(Z^\top W_{t,H} Z)
&= \underbrace{\sum_i (W_{t,H})_{ii}
    \|Z_{i,:}\|_2^2}_{D_{t,H}} \\
&\quad + \underbrace{\sum_{i\neq j}
    (W_{t,H})_{ij}
    \langle Z_{i,:},Z_{j,:}\rangle}_{C_{t,H}}.
\end{aligned}
\end{equation}
We quantify the signed contribution of cross-row terms as
$\mathbb{E}[C_{t,H}]/\mathbb{E}[C_{t,H}+D_{t,H}]$
over different decoding horizons $H$.

\begin{table}[t]
\centering
\caption{Contribution of cross-row interactions to
cumulative squared readout error. Values report the
signed cross-row share of the full quadratic form.}
\label{tab:cross_row}
\small
\begin{tabular}{@{}llccc@{}}
\toprule
Model & State & $H=16$ & $H=128$ & $H=1024$ \\
\midrule
Qwen & STEPQuant@6 & -0.075 & -0.033 & -0.003 \\
Qwen & STEPQuant@4 &  0.053 & 0.036  &  0.028 \\
\midrule
KDA & STEPQuant@6  &  0.007 & -0.004 & -0.014 \\
KDA & STEPQuant@4  &  0.068 &  0.040 &  0.031 \\
\bottomrule
\end{tabular}
\end{table}

As shown in Table~\ref{tab:cross_row}, the net cross-row
contribution under STEPQuant is at most 3.1\% in magnitude
at $H=1024$ across both models and bit widths.
These results support the use of row-wise importance as
a practical approximation to the full readout-error objective.

\subsection{Calibration Statistics for Temporal and Spatial Components}
\label{app:calibration_statistics}
STEPQuant's temporal component uses Lifetime-aware Bit Allocation and
sparse FP16 pivots, while its spatial component uses Key-Row-Aware
Dual-axis Fitting. The statistics below connect their calibration.

\paragraph{Mean log retention.}
For allocation unit $u$, define
\begin{equation}
 \ell_u=\mathbb E_{\mathrm{cal}}[\log r_{t,u}],
 \qquad
 L_u=\sum_{j=0}^{H-1}\exp(2j\ell_u).
 \label{eq:calibration_lifetime}
\end{equation}
The expectation is taken over calibration tokens. For Qwen, $u$ is a
state head and $r_{t,u}=\alpha_{t,u}$; for KDA, $u$ is a key row and
$r_{t,u}=d_{t,u}$. The lifetime weight $L_u$ is applied to the
candidate-format distortion after that distortion is averaged over
state samples.

Let $\mathcal C$ contain the sampled reference states, with each batch
element counted as a separate state sample. Let $S_{s,u}$ be the reference
state of unit $u$ at sample $s$, and $\widehat S^{(b)}_{s,u}$ its
candidate-format reconstruction at precision $b$.

\paragraph{Qwen reconstruction distortion.}
For Qwen, $u$ is an entire head and $n_u=d_kd_v$. Its candidate-format
calibration uses the spatial component's row-impact factors $w_{u,i}$:
\begin{equation}
 d_u(b)=\frac{1}{|\mathcal C|\,n_u}
 \sum_{s\in\mathcal C}\sum_{i,j}
 w_{u,i}^2
 \left(\widehat S^{(b)}_{s,u,ij}-S_{s,u,ij}\right)^2.
 \label{eq:qwen_calibration_distortion}
\end{equation}
The squared row-impact factors $w_{u,i}^2$ weight the reconstruction error.
Weighted squared error is averaged over the elements
within each head and then over state samples; the temporal lifetime
weight $L_u$ is applied afterward.

\paragraph{KDA reconstruction distortion.}
For a KDA key row, $n_u=d_v$ and the calibrated distortion is the
unweighted per-element MSE:
\begin{equation}
 d_u(b)=\frac{1}{|\mathcal C|\,n_u}
 \sum_{s\in\mathcal C}
 \left\|\widehat S^{(b)}_{s,u}-S_{s,u}\right\|_2^2.
 \label{eq:kda_calibration_distortion}
\end{equation}
KDA samples states every eight recurrent updates, averages squared error
over batch elements and value coordinates, and then averages over the
sampled update positions. Once FP16 pivots are fixed, their rows do not
participate in fitting the shared column scales for integer rows.

Thus, Qwen's temporal allocation uses a distortion measure that already
accounts for spatial row impact. In KDA, the allocation MSE is unweighted,
but the candidate reconstruction is produced with the spatial codec and
the shared column fit excludes FP16 pivot rows. The two components have
distinct roles, while sharing the calibrated state representation.
Within each model, all allocation units have the same $n_u$. Converting
per-element MSE to total squared error therefore multiplies the
allocation objective by a model-specific constant and does not change
the selected precision assignment.

\section{Analysis of Temporal Error Accumulation and Spatial Row Impact}
\label{app:diagnostics}
These controlled experiments support the temporal analysis in
Section~\ref{sec:analysis} and the spatial analysis in
Sections~\ref{sec:key-row} and~\ref{sec:two-axis}. They use reference
input streams to distinguish error persistence from differences in
key-row impact and state geometry.

\subsection{Effect of Delta Feedback on Quantization Error}
\label{app:oracle}
For GDN, the directional contraction is
\begin{equation}
 \|\alpha(I-\beta kk^\top)E\|_F^2
 =\alpha^2[\|E\|_F^2-\beta(2-\beta\|k\|_2^2)\|k^\top E\|_2^2].
 \label{eq:self_correction}
\end{equation}
Let $P=I-\beta kk^\top$. Expanding $P^\top P$ gives
$I-\beta(2-\beta\|k\|_2^2)kk^\top$, which proves
Equation~\ref{eq:self_correction}. For KDA, the corresponding subtraction
is applied to $D_tE_{t-1}$, retaining the native order of decay and correction.
The mechanism does not require the state transition to amplify perturbations:
repeated injections can accumulate under a non-expansive transition.

The exact-read oracle replaces only the state inside the Delta residual:
\[
 \widetilde S_t^{\rm oracle}
 =D_t\widehat S_{t-1}^{\rm oracle}
 +\beta_t k_t(v_t^\top-k_t^\top D_tS_{t-1}).
\]
Subtracting the reference update leaves $D_tE_{t-1}^{\rm oracle}$.
This intervention removes the correction of the path's own state error.
With continuous quantization it also changes future rounding errors. Therefore we use a matched single-injection experiment to isolate feedback.

Each of four fixed C4 \citep{raffel2020exploring} Qwen trajectories provides native
keys, values, queries, and gates for 8,192 updates. We use the deterministic Cartesian subset of
global model-layer indices $\{0,8,16,24,32,40,48,62\}$ and global heads
$\{0,12,24,36\}$, giving 32 heads fixed before replay and independent of
quantization outcomes. We inject one quantization error at step 256 and
perform the remaining 7,936 updates without further
quantization. Both paths start from the identical error. We sum squared output error
over heads, trajectories, and subsequent steps. Relative to native Delta
feedback, the exact-read oracle increases this error by $26.82\times$ for
INT6 and $18.65\times$ for INT8; both ratios exceed one on each of the four
trajectories.

\subsection{Prefill and Decode Prediction Drift}
\label{app:decode_probe}
An uninterrupted native prefill evolves an uncompressed state and packs it
once at the boundary. Decode then reads and rewrites the compressed state
after each token. This probe measures prediction drift caused by repeated
compressed-state updates.

On four held-out C4 streams per model, each with a 2,048-token native prefill
and 6,144 forced decode updates, prefill predictions match the FP32-state
reference. Mean excess NLL (quantized minus reference) over the first and
last 256 compressed-read predictions rises from 0.102 to 2.102 for Qwen
INT6, and from 0.0228 to 0.1422 for KDA INT6. Across all 6,144 decode
predictions, STEPQuant@6bit has mean excess NLL of 0.0011 on Qwen and
$-0.0120$ on KDA.

\subsection{Readout Error at Fixed Precision}
\label{app:same_gate}
For Qwen, the calibrated row-impact factor satisfies
$w_i\propto\omega_i^{1/8}$, where
$\omega_i=\mathbb E[(A_t^\top q_t)_i^2]$ includes the immediate Delta
transition and readout (Section~\ref{sec:fitting}).
Within a head, the normalization constant cancels
in row-impact ratios. We recover a proportional row-impact score as $w_i^8$
from the frozen full-precision calibration record, retaining its numerical
floor, and compute the interpolated P90/P10 ratio across 128 key channels.
Across all 2,304 heads, the median ratio is 15.95. The 10th and 90th
percentiles across heads are 3.74 and 79.84. All channels within each such
head share exactly the same scalar gate. This difference is thus additional
to gate lifetime and directly motivates the spatial component's
within-head row-impact factor.

Figure~\ref{fig:lifetime_state}d isolates row impact at fixed assigned
precision. A group is one integer Qwen head, or the integer rows sharing
both head and bit width in KDA, using frozen STEPQuant@6bit assignments.
FP16 pivots are excluded. Requiring at least eight rows per group retains
2,272 Qwen groups (290,816 rows) and 1,567 KDA groups (80,333 rows).
1,075 KDA integer rows in smaller groups are excluded, without filtering
on measured error. Qwen rows in a group share a scalar gate. KDA rows
can have different keywise gates despite sharing their assigned bit width.

On reference-state traces, we inject a unit-Frobenius-norm perturbation
$E_i$ confined to key row $i$. With
$A_t=(I-\beta_t k_tk_t^\top)D_t$ and $g_t=A_t^\top q_t$, its isolated
next-read squared error is exactly $\|E_i^\top g_t\|_2^2=g_{t,i}^2$.
We average 32 probe positions following 128 replay updates over four
Qwen and two KDA C4 streams. Direct propagation checks agree with this
identity within $7.1\times10^{-7}$ relative error.

Rows are ranked within each group using independent WikiText calibration:
frozen $w_i^8$ for Qwen and one-step transported-query row-impact scores
for KDA.
Each row's held-out read error is divided by its own group's mean.
We split each ranked group into eight near-equal bins and average bin
means with equal group weight. Points are measured means at the averaged
percentile centers. Curves use shape-preserving interpolation without
extrapolation. The highest-to-lowest bin ratios are 117.4 for Qwen and
35.1 for KDA. These within-group contrasts motivate Key-Row-Aware
Dual-axis Fitting even after lifetime-aware precision allocation.

\subsection{Key-Row Quantization and Perplexity}
\label{app:row_impact_ppl}
The perplexity experiment in Figure~\ref{fig:read_sensitivity}(a) tests
whether calibrated row impact predicts the effect of state quantization
on model outputs (Section~\ref{sec:key-row}). Within each head, we rank
key rows by the independently calibrated $\omega_i$ and divide them into
eight near-equal groups. We quantize one ranked group at a time to INT4
across the model, keeping all other rows in full precision. Perplexity
is measured on eight 1,024-token streams per model and compared with
the FP32-state reference. Markers show the measured bin results, and
curves use shape-preserving interpolation. Higher-impact groups generally
cause larger perplexity increases in both models. Unlike the equal-norm
probe above, this experiment measures the effect of quantizing actual
state values on the model's predictive distribution.

\subsection{Gate Half-Life and Accumulated INT6 State Error}
\label{app:lifetime_state}
\begin{figure}[!htb]
 \centering\includegraphics[width=\linewidth]{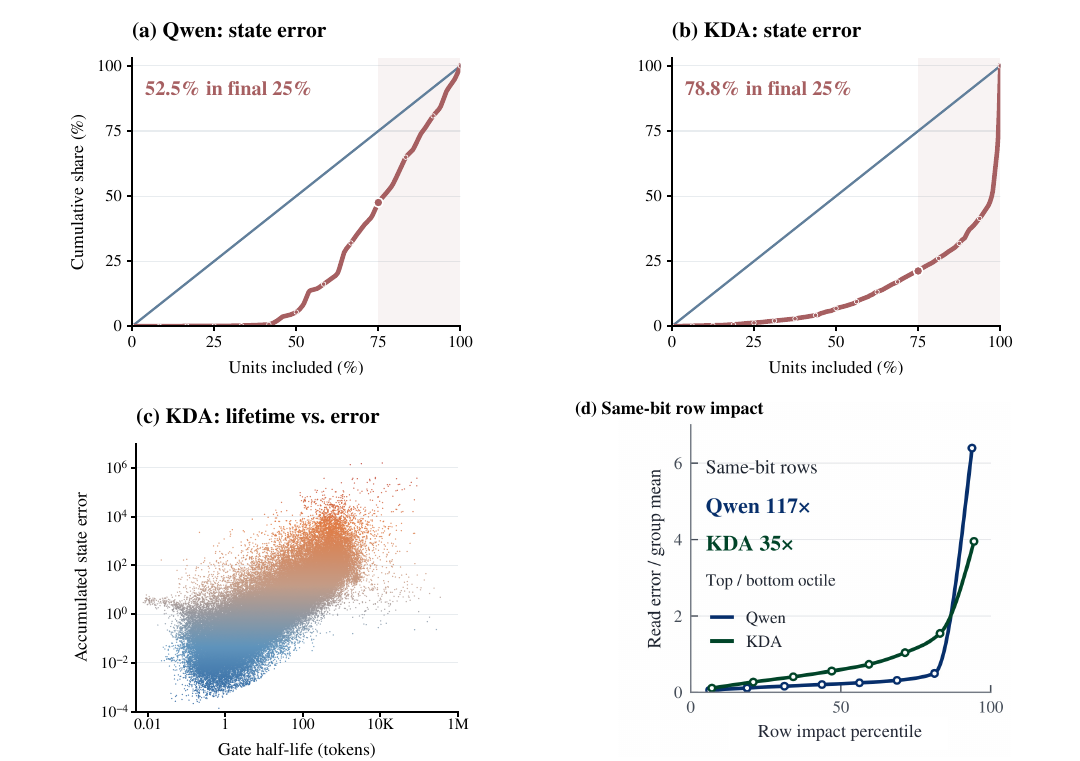}
\caption{\textbf{Gate lifetime and key-row impact of recurrent state.}
 (a--b) Cumulative INT6 squared error (red) and unit count (blue), ordered
 by increasing gate half-life within each layer or head. The longest-lived quarter accounts for
 52.5\% of Qwen error and 78.8\% of KDA error across all 2,304 heads and
 81,920 channels. (c) KDA half-life versus accumulated INT6 error for
 79,864 channels after excluding near-zero errors; color encodes error.
 (d) Held-out equal-norm readout-error contrast across row-impact
 octiles within fixed-precision groups, using independent WikiText ranking.}
 \label{fig:lifetime_state}
\end{figure}

Figure~\ref{fig:state_capacity}b shows the Qwen heads, and
Figure~\ref{fig:lifetime_state}c shows the corresponding KDA key channels.
Both scatters measure state error under uniform INT6.
Full-population Spearman correlations are 0.8004 and 0.8017, respectively.
For legibility, the KDA scatter omits 2,056 channels (2.51\%) with
squared error below $10^{-15}$.
The remaining 79,864 channels are shown on logarithmic axes.
A Qwen head has 16,384 state entries and a KDA key row has 128, so raw
error magnitudes across models are not directly comparable.

For Figure~\ref{fig:lifetime_state}, frozen WikiText calibration estimates
$\ell_u$ as defined in Appendix~\ref{app:calibration_statistics}, and gives the gate
half-life $\tau_u=\log(2)/(-\ell_u)$. This gate-only statistic describes
forgetting. The Delta correction remains part of the measured recurrence.
The error experiment uses four frozen C4 sequences per model with their
model-specific tokenizers. After a two-token FP32 boundary, a separate INT6
trajectory follows the reference keys, values, and gates for 2,046 decode
updates. Every key row uses an absmax/31 scale stored in FP16, clamped to
$[2^{-14},65504]$, and nearest-integer codes in $[-31,31]$. Quantization is
applied at the boundary and after every recurrent update. For unit $u$ we
record
\[
 D_u=\sum_{r=1}^{4}\sum_{t=1}^{2046}
 \norm{(\widehat S_{r,t}-S_{r,t})_u}_F^2.
\]
A Qwen unit is a whole head matrix. A KDA unit is a key row containing 128
value components. These units partition each state, so their squared errors
sum exactly. We sort by lifetime within each group, sum $D_u$ across groups at each rank, and then accumulate from shortest
to longest. The error curve is normalized by its model-wide total. No per-layer or
per-head normalization precedes pooling. The longest quarter contains
12 heads per Qwen layer or 32 channels per KDA head. Shape-preserving cubic
interpolation passes through every measured cumulative-rank point.

Qwen replays captured native FP32 inputs, checking reference readouts against
the capture. KDA executes its native recurrent kernel on a separate shadow
state. These controlled mechanism replays isolate persistent-state
distortion along reference input streams.

\subsection{Lifetime Rankings and Dual-Axis State Geometry}
\label{app:dl_motivation}
\begin{figure}[!htbp]
 \centering\includegraphics[width=\linewidth]{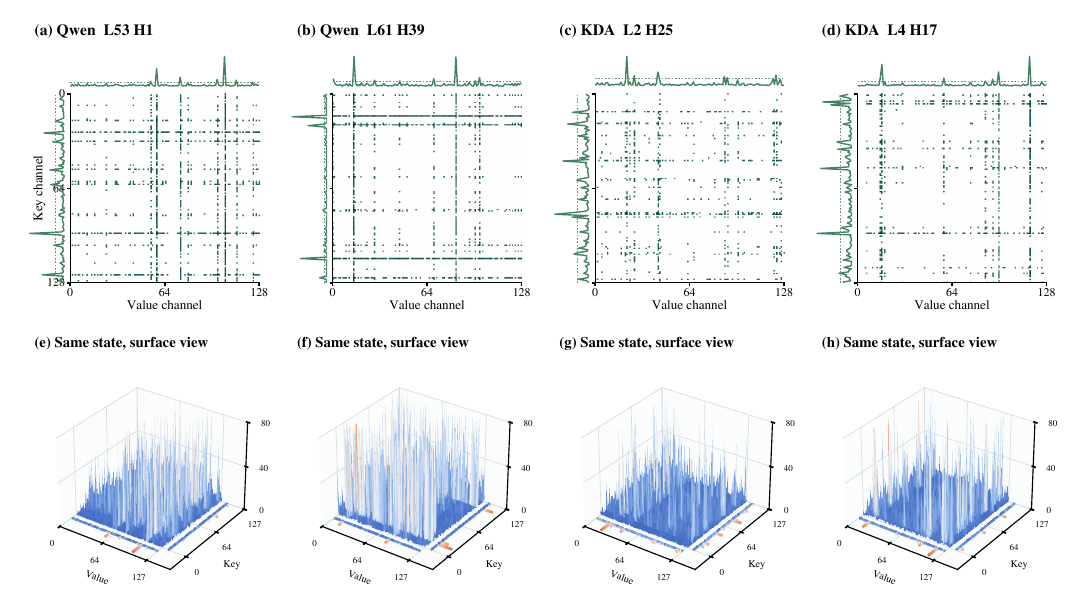}
 \caption{\textbf{Dual-axis state geometry across Qwen and KDA.}
 Four native-order $128\times128$ states (Qwen, Qwen, KDA, KDA).
 Top: entries below $10\times$ the matrix-median absolute magnitude are white;
 larger entries use graded greens. Marginal traces show row and column RMS
 relative to their axis medians. Bottom: matching surfaces use the same
 style as the Qwen surface in Figure~\ref{fig:read_sensitivity}(b).
 Floor traces show the same normalized
 RMS profiles, rescaled for display; heights are capped at $80\times$.
 Each state has seven or eight
 rows and columns above $5\times$ their axis-median RMS.}
 \label{fig:dual_axis_gallery}
\end{figure}

\begin{figure}[!htbp]
 \centering\includegraphics[width=.95\linewidth]{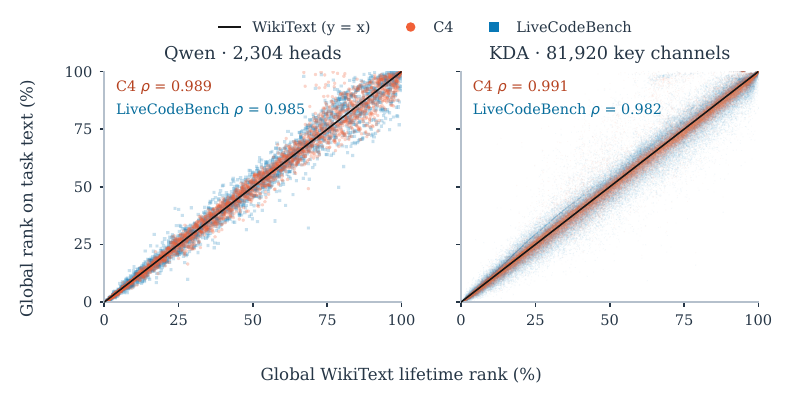}
 \caption{\textbf{Global lifetime rankings remain similar across text sources.}
 Qwen (2,304 heads) and KDA (81,920 key channels) are ranked separately,
 from shortest to longest gate half-life. Each point compares its global
 WikiText rank percentile with its C4 (orange-red) or LiveCodeBench
 (sky blue) percentile. The thin black diagonal is WikiText compared
 with itself ($y=x$).
 Spearman coefficients use the complete population in each panel.}
 \label{fig:lifetime_transfer}
\end{figure}

Figure~\ref{fig:lifetime_transfer} compares gate-lifetime ordering under
WikiText-2 \citep{merity2016pointer}, C4, and LiveCodeBench
\citep{jain2025livecodebench} text. The frozen WikiText calibration is
independent of the task streams. The BF16 Qwen gate observer covers all 2,304 recurrent
heads using four archived 2,048-token C4 streams and the first four unique
sample-0 LiveCodeBench trajectories in sorted archive order. The latter
combine original prompts and FP32-reference generated token IDs, truncated
at 2,048 tokens without padding, for 6,218 observed tokens per head.
KDA covers all 81,920 key channels. Its C4 inputs are four 2,048-token
evaluation streams. Its LiveCodeBench inputs follow the same selection rule,
combining original prompts with archived KDA W4 FP32-state generated tokens;
the BF16 KDA gate observer processes 8,192 C4 and 3,275 LiveCodeBench
tokens per layer.

For each text source, we compute token-weighted mean log gate retention
and rank the resulting half-lives from shortest to longest across all units
of each model. Figure~\ref{fig:lifetime_transfer} plots these global ranks
as percentiles. Full-population Spearman correlations with WikiText are
0.989 (C4) and 0.985 (LiveCodeBench) for Qwen, and 0.991 and 0.982 for
KDA. For Qwen, the short/middle/long classes defined by within-layer ranks
1--12, 13--36, and 37--48 retain their WikiText labels for 93.6\% of heads
on C4 and 90.1\% on LiveCodeBench.

Figure~\ref{fig:dual_axis_gallery} uses two Qwen C4 snapshots at step
2,048 from distinct layers and heads. The two KDA snapshots are replayed
from captured C4 keys, values, gates, and update coefficients at steps 128
or 256, also from distinct layers and heads. The native channel order and
all matrix entries are retained. We chose the four examples by the $5\times$
RMS criterion in the caption. Across the full KDA capture of 3,840
snapshots at steps 128, 256, and 512, median maximum-to-median RMS
contrasts are $12.7\times$ for key rows and $5.4\times$ for value columns;
both contrasts exceed $3\times$ in 3,262 snapshots (84.9\%).

For the Qwen state-geometry analysis in Section~\ref{sec:two-axis},
FP32 state matrices are reconstructed
from native recorded inputs at positions 128, 512, and 2,048. Define
$r_i^{\rm rms}=\sqrt{\frac1{128}\sum_j S_{ij}^2}$ and
$c_j^{\rm rms}=\sqrt{\frac1{128}\sum_i S_{ij}^2}$. The outlier contrasts
are $\max_i r_i^{\rm rms}/\operatorname{median}_i r_i^{\rm rms}$ and the
analogous column ratio. Medians use linear interpolation. Four C4 streams
provide 27,648 matrices and two AIME streams provide 13,824. These repeated
snapshots describe state geometry and are not independent model replicates.
On C4, the median contrasts are $10.3\times$ for key rows and
$19.4\times$ for value columns, and both exceed $3\times$ in 98.6\%
of snapshots, as reported in Section~\ref{sec:two-axis}.
The representative Qwen surface in Figure~\ref{fig:read_sensitivity}(b)
uses global layer 57, head 2, C4 stream 2
at step 2,048. It was selected for clearly visible extended row and column
ridges whose intersection is also an outlier. All $128\times128$ entries
retain their native order. The linear height axis ends at 0.2, with larger
values clipped for display.

The time-resolved row and column profiles in Figure~\ref{fig:read_sensitivity}(c)
apply the same RMS normalization at each recorded decode update.
They show how large-magnitude channels persist over the trajectory;
the snapshot-based population statistics use all recorded matrices.

\FloatBarrier
\section{Calibration, Precision Allocation, and Decode Procedure}
\label{app:method_details}
This appendix complements STEPQuant's Lifetime-aware Bit Allocation,
Key-Row-Aware Dual-axis Fitting, and execution steps in
Sections~\ref{sec:lmsq} and~\ref{sec:fitting} and
Appendix~\ref{app:algorithm}.

\subsection{Row-Impact Calibration for KDA}
For KDA, the spatial component uses the same one-step transported-read definition as Qwen,
evaluated in KDA's native state coordinates. Write
$D_t=\diag(d_{t,1},\ldots,d_{t,d_k})$. Since
$A_t=(I-\beta_tk_tk_t^\top)D_t$, the coordinate read sensitivity is
\begin{equation}
 g_{t,i}=(A_t^\top q_t)_i
 =d_{t,i}\left[q_{t,i}-\beta_t k_{t,i}(k_t^\top q_t)\right],
 \qquad
 \omega_i^{\rm KDA}=\E_{\rm cal}[g_{t,i}^2].
 \label{eq:kda_sensitivity}
\end{equation}
We normalize this profile in the same way as for Qwen:
\begin{equation}
 w_i=\frac{(\omega_i^{\rm KDA})^{\gamma/2}}
            {\gm_j\bigl((\omega_j^{\rm KDA})^{\gamma/2}\bigr)},
 \qquad \gamma=0.25.
 \label{eq:kda_precond}
\end{equation}
The resulting factors enter the same row-scale rule
$r_i=m_i^{1/2}w_i^{-1/2}$ and weighted column fit used for Qwen.
The column-fitting objective weights squared reconstruction errors by
$w_i^2$. KDA's post-convolution keys,
queries, and gates are measured in native coordinates.

\paragraph{Code normalization and shared fitting.}
\label{app:shared_fitting}
For KDA, $\mathcal C_b=\{2^{8-b}q:q\in\mathbb Z,\ |q|\leq2^{b-1}-1\}$.
Only the $b$-bit integer $q_{ij}$ is stored; reconstruction and fitting use
$z_{ij}=2^{8-b_i}q_{ij}$. Qwen's 4/6/8-bit codebooks
use the same integer bounds without this fixed normalization.
Let $\mathcal I$ denote the non-pivot integer rows in a head.
Writing $v_{ij}=r_iz_{ij}$, the fixed-code update is
$c_j=\sum_{i\in\mathcal I}w_i^2 v_{ij}X_{ij}/\sum_{i\in\mathcal I}w_i^2 v_{ij}^2$,
with a positive numerical floor. A zero denominator retains the previous scale.

\subsection{Calibration Data, Precision Candidates, and FP16 Pivots}
\label{app:calibration}
Both models, at both precision budgets and with both weight formats, use
32 WikiText-2 training segments of 2,048 tokens each.
\begin{table}[!htbp]
 \centering\small
 \caption{\textbf{Model-specific STEPQuant configuration.} 
    FP16 pivot and scale-metadata costs are accounted for separately from the nominal bit budget.
 }
 \begin{tabular}{lcc}
 \toprule
 Property & Qwen3.8-27B & Kimi-Linear-48B-A3B \\
 \midrule
 Recurrent layers & 48 & 20 \\
 State heads per recurrent layer & 48 & 32 \\
 Allocation unit & entire head & key row \\
 Number of allocation units & 2,304 & 81,920 \\
 Integer candidates (@4 / @6) & $\{2,4,6,8\}$ / $\{4,6,8\}$ & $\{2,4,6,8\}$ / $\{4,6,8\}$ \\
 Optimizer & multiple-choice DP & Lagrangian allocation \\
 FP16 pivots & 32 heads & 512 key rows \\
 Pivot ranking & residual allocation risk & distortion-reduction risk \\
 \bottomrule
 \end{tabular}
\label{tab:config}
\end{table}
\paragraph{Qwen calibration.}
The row-impact-weighted MSE in
Equation~\ref{eq:qwen_calibration_distortion} supplies the head-level
distortion term for Lifetime-aware Bit Allocation. The spatial component
fits and writes the serving state in native coordinates.

\paragraph{KDA calibration.}
KDA combines gate-derived lifetime with the per-row candidate-format MSE
in Equation~\ref{eq:kda_calibration_distortion}. Its calibrated row-impact
scores supply the spatial component's preconditioning factors.
Pivot selection scores the lifetime-weighted reduction from
integer reconstruction error to FP16 reconstruction error. With pivots
fixed, candidate distortions are calibrated excluding those rows from
the shared integer column fit, and remaining rows are allocated under the integer
budget. Candidate distortions are frozen for discrete allocation.
Online Key-Row-Aware Dual-axis Fitting subsequently refits the shared
scales to each updated state.

\paragraph{Allocation optimizers.}
Qwen uses a multiple-choice dynamic program over candidate precisions and
remaining budget. KDA minimizes the Lagrangian per unit for a shared budget
multiplier, then repairs the discrete assignment to meet the integer budget.
The precision menu and budget change between four-bit and six-bit settings,
while the lifetime-weighted objective remains the same.

\subsection{Recurrent Update and Quantized Writeback}
\label{app:algorithm}
The codebooks and fixed-code column-scale update are specified in
Appendix~\ref{app:shared_fitting}. The steps below combine this fitting
with the recurrent update and packed-state storage.
\begin{table}[!htbp]
\centering
\caption{\textbf{One persistent STEPQuant decode step.} The same logical
workflow supports four-bit and six-bit budgets.}
\begin{tabular}{p{.94\linewidth}}
\toprule
\textbf{Inputs:} compressed state. $q_t,k_t,v_t,D_t,\beta_t$. Frozen
precision map, pivot mask, and row-impact scores.\\[3pt]
1. Reconstruct the previous state from integer codes and scales, or from
FP16 values for pivot units.\\
2. Compute $X=D_t\widehat S_{t-1}
+\beta_t k_t(v_t^\top-k_t^\top D_t\widehat S_{t-1})$.\\
3. Emit $\widehat y_t=X^\top q_t$ before requantization and continue
subsequent model computation.\\
4. Derive row factors using Key-Row-Aware Dual-axis Fitting
(Section~\ref{sec:fitting}). Qwen's two-bit
path shares row factors within value groups.\\
5. For KDA, jointly fit one shared column-scale vector with squared row-impact factors over all
non-pivot integer rows within each head. Qwen fits each head at
its assigned precision. Its two-bit path uses signed magnitude levels.\\
6. Pack integer codes and scales. Store pivot values in FP16.
Keep the old representation valid until its readers finish.\\
7. Complete writeback before the next recurrent step uses the new page.\\[3pt]
\textbf{Persistent output:} updated packed codes, scales, and FP16 pivots.\\
\bottomrule
\end{tabular}
\end{table}

\FloatBarrier
\section{Storage Cost of Codes, Scales, and FP16 Pivots}
\label{app:capacity}
Let $N=L_h n_h d_kd_v$ denote all recurrent-state elements of one request.
Qwen has $N=48\cdot48\cdot128^2=37,748,736$. KDA has
$N=20\cdot32\cdot128^2=10,485,760$. The FP32 recurrent-state
representation is $4N$ bytes. Full-attention KV caches, convolution state,
model weights, shared metadata, temporary workspaces, and allocator overhead
are separate from this representation.

\paragraph{Compact dual-axis representation.}
For a head storing one FP16 row vector and one FP16 column vector,
these two vectors contribute
\[
 \frac{16(d_k+d_v)}{d_kd_v}=0.25
 \quad\text{bits per element},\qquad d_k=d_v=128.
\]
In Qwen's six-bit configuration, 32 pivots replace eight-bit heads with FP16.
Including two scale vectors for every head gives
\begin{equation}
 b_{\rm Qwen}=6+\frac{16}{128}+\frac{16}{128}
                  +\frac{32}{2304}(16-8)=6.3611.
 \label{eq:compact_qwen}
\end{equation}
This analytical count reserves two scale slots for every head, including
pivots; the packed serving layout omits pivot scale slots.

For the Qwen four-bit configuration, the evaluated shared-scale recurrent-state representation
contains 21,804,032 bytes per request, including FP16 scales and 32 FP16
pivot heads. Its compact cost is 4.6209 bits/value, giving a
$150,994,944/21,804,032=6.9251$ recurrent-state representation ratio
relative to FP32.
Figure~\ref{fig:state_capacity} rounds this ratio to $6.93\times$.
Compact counts exclude tensor-parallel padding and allocator overhead.

KDA uses row-level allocation with one shared FP16 column-scale vector
per head at both budgets. Codebook normalization is fixed by precision
and consumes no metadata. Thus one row vector and one column vector cost
$16/128+16/128=0.25$ bits/value, independent of the number of active
precisions (three at six bits, up to four at four bits).
For reporting, we use the nominal pre-pivot code budget $\bar b$ and
count 512 FP16 pivot rows as replacements for eight-bit rows at both budgets.
Thus the residual integer budget is $\bar b N-8N_{\rm piv}$, where
$N_{\rm piv}=512d_v$. Under this accounting convention, the compact count is
\begin{equation}
 b_{\rm KDA}(\bar b)=\bar b+\frac{16}{128}
               +\frac{16}{128}
               +\frac{512}{81920}(16-8)=\bar b+0.30.
 \label{eq:compact_kda}
\end{equation}
This gives 4.30 and 6.30 bits/value at the four- and six-bit budgets,
respectively, for both weight formats. The corresponding per-request
sizes are 5.375 and 7.875\,MiB, versus 40\,MiB for FP32, giving
$7.44\times$ and $5.08\times$ compression. These are analytical compact-format
counts at the nominal budget, rather than measured allocation sizes. Tensor-parallel page padding and allocator overhead are excluded.

\begin{table}[!htbp]
 \centering\small\setlength{\tabcolsep}{8pt}
 \caption{\textbf{Compact recurrent-state representation (bits/value).}
 Integer codes, FP16 scales, and pivot replacement are included.
 Qwen@4 uses the packed byte count above; the other entries follow
 Equations~\ref{eq:compact_qwen} and~\ref{eq:compact_kda}.}
 \begin{tabular}{llrr}
 \toprule
 Model & Weights & STEPQuant@6bit & STEPQuant@4bit \\
 \midrule
 Qwen & BF16 & 6.361 & 4.621 \\
 Qwen & W4A16 & 6.361 & 4.621 \\
 KDA & BF16 & 6.300 & 4.300 \\
 KDA & W4A16 & 6.300 & 4.300 \\
 \bottomrule
 \end{tabular}
 \label{tab:storage}
\end{table}

\section{Per-Task Accuracy and Generation Length}
\label{app:provenance}

\subsection{Scoring Generated Answers on Short Tasks}
\label{app:short_scoring}
The six short tasks use greedy generation and score the answer extracted
from the generated response, rather than ranking candidate likelihoods.
A response that does not provide an extractable answer is counted as
incorrect. WinoGrande has two choices, yet uniform INT4 scores 46.57\%
on Qwen and 17.36\% on KDA (Table~\ref{tab:short}). At this precision,
some generations lose the requested answer format and do not supply a
valid choice. Thus accuracy can fall below 50\% even on a binary task:
the score also captures the model's ability to follow the instruction
and produce an answer during decode.

\subsection{Generation length with BF16 weights}
\label{app:bf16length}
Table~\ref{tab:length} reports the task-level generation lengths underlying
Figure~\ref{fig:serving_throughput}a--b. Each entry averages all evaluated
outputs for that task, including thinking, incorrect answers, and capped
outputs. The final column averages the seven task means before rounding.
\begin{table}[!htbp]
 \centering\small\setlength{\tabcolsep}{3pt}
 \caption{\textbf{Mean generated tokens with BF16 weights (thousands).}}
 \begin{tabular}{@{}lrrrrrrrr@{}}
\toprule
State & LCB v6 & EvalPlus & AIME 26 & MATH-500 & HMMT & GPQA-D & IFBench & Avg. \\
\midrule
\multicolumn{9}{@{}l}{\textit{Qwen3.8-27B}} \\
\addlinespace[2pt]
FP32 & 5.64 & 0.62 & 9.73 & 1.67 & 18.15 & 5.10 & 4.71 & 6.52 \\
INT8 & 9.74 & 0.87 & 15.44 & 2.32 & 29.95 & 6.04 & 9.02 & 10.48 \\
INT6 & 19.63 & 1.43 & 25.62 & 5.68 & 33.24 & 11.31 & 16.72 & 16.23 \\
INT4 & 29.51 & 2.17 & 53.28 & 36.30 & 50.74 & 48.62 & 53.53 & 39.16 \\
\rowcolor{methodwash}
STEPQuant@6bit & \textbf{4.44} & \textbf{0.64} & \textbf{8.51} & \textbf{1.62} & \textbf{20.11} & \textbf{5.12} & \textbf{5.89} & \textbf{6.62} \\
\rowcolor{methodwash}
STEPQuant@4bit & \textbf{7.46} & \textbf{0.69} & \textbf{13.01} & \textbf{1.65} & \textbf{20.55} & \textbf{4.95} & \textbf{7.39} & \textbf{7.96} \\
\midrule
\multicolumn{9}{@{}l}{\textit{Kimi-Linear-48B-A3B-Instruct}} \\
\addlinespace[2pt]
FP32 & 9.47 & 1.10 & 18.89 & 4.17 & 33.64 & 9.15 & 6.70 & 11.87 \\
INT8 & 12.86 & 1.94 & 27.55 & 4.20 & 38.86 & 12.41 & 6.10 & 14.84 \\
INT6 & 13.75 & 3.35 & 29.29 & 7.47 & 40.39 & 17.36 & 10.45 & 17.44 \\
INT4 & 22.32 & 15.86 & 63.40 & 37.73 & 64.22 & 57.12 & 26.84 & 41.07 \\
\rowcolor{methodwash}
STEPQuant@6bit & \textbf{9.95} & \textbf{1.25} & \textbf{20.11} & \textbf{3.92} & \textbf{31.73} & \textbf{8.64} & \textbf{6.25} & \textbf{11.70} \\
\rowcolor{methodwash}
STEPQuant@4bit & \textbf{9.80} & \textbf{1.27} & \textbf{21.34} & \textbf{3.14} & \textbf{32.32} & \textbf{7.31} & \textbf{5.61} & \textbf{11.54} \\
\bottomrule
\end{tabular}

 \label{tab:length}
\end{table}

\subsection{Generation length with W4A16 weights}
\label{app:w4length}

\begin{table}[!htbp]
 \centering\small\setlength{\tabcolsep}{3.5pt}
 \caption{\textbf{Mean generated tokens with W4A16 weights.}
 Values are thousands of output tokens over all evaluated samples,
 with the same definition and task ordering as Table~\ref{tab:length}.
}
 \begin{tabular}{@{}lrrrrrrrr@{}}
\toprule
State & LCB v6 & EvalPlus & AIME 26 & MATH-500 & HMMT & GPQA-D & IFBench & Avg. \\
\midrule
\multicolumn{9}{@{}l}{\textit{Qwen3.8-27B}} \\
\addlinespace[2pt]
FP32 & 8.99 & 0.91 & 10.87 & 1.66 & 18.00 & 4.97 & 5.11 & 7.22 \\
\rowcolor{methodwash}
STEPQuant@6bit & \textbf{8.77} & \textbf{0.89} & \textbf{11.03} & \textbf{1.67} & \textbf{21.37} & \textbf{5.02} & \textbf{5.19} & \textbf{7.70} \\
\rowcolor{methodwash}
STEPQuant@4bit & \textbf{10.29} & \textbf{0.95} & \textbf{13.56} & \textbf{1.80} & \textbf{23.88} & \textbf{5.25} & \textbf{10.80} & \textbf{9.50} \\
\midrule
\multicolumn{9}{@{}l}{\textit{Kimi-Linear-48B-A3B-Instruct}} \\
\addlinespace[2pt]
FP32 & 9.27 & 1.35 & 24.62 & 4.07 & 34.20 & 8.92 & 6.49 & 12.70 \\
\rowcolor{methodwash}
STEPQuant@6bit & \textbf{9.36} & \textbf{1.38} & \textbf{23.58} & \textbf{4.00} & \textbf{34.65} & \textbf{8.99} & \textbf{6.40} & \textbf{12.62} \\
\rowcolor{methodwash}
STEPQuant@4bit & \textbf{10.30} & \textbf{1.41} & \textbf{25.67} & \textbf{4.37} & \textbf{35.80} & \textbf{9.11} & \textbf{6.19} & \textbf{13.26} \\
\bottomrule
\end{tabular}

 \label{tab:w4length}
\end{table}

At the four-bit budget with W4A16 weights, Qwen's mean accuracy is
0.34 percentage points below FP32, while mean generation length rises
from 7.22K to 9.50K tokens (approximately 31.6\%). KDA's mean accuracy
falls by 2.29 points, with mean length increasing from 12.70K to 13.26K
(approximately 4.4\%).

\subsection{Additional Component Ablation}
\label{app:ablation}

As a supplement to the component ablation in Sec.~\ref{sec:ablation}, Table~\ref{tab:ablation_kimi} reports the corresponding results on Kimi under nominal 4- and 6-bit budgets. Consistent with Qwen, spatial fitting outperforms DSQ, sparse FP16 pivots improve the temporal component, and combining both components achieves further accuracy gains. STEPQuant also consistently outperforms Q-Mamba when using the same bit allocation and FP16 pivots, further validating the effectiveness of our spatial quantization design.

\begin{table}[!htbp]
\centering
\caption{Component ablation on Kimi-Linear-48B-A3B-Instruct.
Accuracy (\%) with BF16 weights. Avg. denotes the average
across three long-generation benchmarks.}
\label{tab:ablation_kimi}
\begin{minipage}[t]{0.49\linewidth}
\centering\scriptsize\setlength{\tabcolsep}{4pt}
\renewcommand{\arraystretch}{0.95}
\textbf{Nominal 6-bit budget}\par\vspace{2pt}
\begin{tabular}{@{}lrrrr@{}}
\toprule
Variant & AIME & GPQA & LCB & Avg. \\
\midrule
FP32 & 68.33 & 69.70 & 54.52 & 64.18 \\
INT6 & 22.71 & 57.83 & 42.37 & 40.97 \\
Q-Mamba@6bit & 38.54 & 57.07 & 44.55 & 46.72 \\
Q-Mamba + alloc./pivots & 61.88 & 67.93 & 52.80 & 60.87 \\
Spatial only & 63.54 & 66.67 & 52.80 & 61.00 \\
Temporal w/o pivots & 41.46 & 59.85 & 48.87 & 50.06 \\
Temporal only & 61.67 & 66.16 & 48.91 & 58.91 \\
\rowcolor{methodwash}
STEPQuant@6bit & \textbf{67.71} & \textbf{68.43} & \textbf{54.86} & \textbf{63.67} \\
\bottomrule
\end{tabular}
\end{minipage}\hfill
\begin{minipage}[t]{0.49\linewidth}
\centering\scriptsize\setlength{\tabcolsep}{4pt}
\renewcommand{\arraystretch}{0.95}
\textbf{Nominal 4-bit budget}\par\vspace{2pt}
\begin{tabular}{@{}lrrrr@{}}
\toprule
Variant & AIME & GPQA & LCB & Avg. \\
\midrule
FP32 & 68.33 & 69.70 & 54.52 & 64.18 \\
INT4 & 0.00 & 14.58 & 28.82 & 14.47 \\
Q-Mamba@4bit & 0.00 & 10.10 & 28.91 & 13.00 \\
Q-Mamba + alloc./pivots & 8.13 & 41.92 & 41.00 & 30.35 \\
Spatial only & 26.67 & 50.51 & 39.51 & 38.89 \\
Temporal w/o pivots & 4.58 & 35.35 & 33.74 & 24.56 \\
Temporal only & 8.33 & 40.91 & 38.01 & 29.08 \\
\rowcolor{methodwash}
STEPQuant@4bit & \textbf{59.17} & \textbf{64.14} & \textbf{53.29} & \textbf{58.87} \\
\bottomrule
\end{tabular}
\end{minipage}

\end{table}

\paragraph{Spatial Design Ablation.}
We further examine the contributions of dual-axis scaling and
row-impact weighting. First, we compare row-only, column-only,
and dual-axis scaling under uniform 6-bit quantization, without
mixed-precision allocation or FP16 pivots. We then isolate the
effect of row-impact weighting by setting $\omega_i=1$ while
keeping the dual-axis fitting procedure unchanged, under both
4-bit and 6-bit budgets.

\begin{table}[!htbp]
\centering
\caption{Spatial design ablation on three long-reasoning tasks
with BF16 weights. All variants use uniform precision without
mixed-precision allocation or FP16 pivots.}
\label{tab:spatial_design_ablation}
\small
\setlength{\tabcolsep}{5pt}
\begin{tabular}{@{}lclrrrr@{}}
\toprule
Model & Bits & Method & AIME & GPQA-D & LCB & Avg. \\
\midrule
Qwen & 6 & Row-only & 33.96 & 51.52 & 30.43 & 38.63 \\
     & 6 & Column-only & 45.21 & 57.32 & 34.03 & 45.52 \\
     & 6 & Dual-axis ($\omega=1$) & 77.08 & 79.04 & 79.15 & 78.42 \\
     & 6 & Dual-axis (with $\omega$) & 79.79 & 80.68 & 80.09 & 80.19 \\
\cmidrule(lr){2-7}
     & 4 & Dual-axis ($\omega=1$) & 67.92 & 62.12 & 67.96 & 66.00 \\
     & 4 & Dual-axis (with $\omega$) & 75.21 & 70.71 & 75.92 & 73.95 \\
\midrule
Kimi & 6 & Row-only & 22.71 & 57.83 & 42.37 & 40.97 \\
     & 6 & Column-only & 28.96 & 55.56 & 43.22 & 42.58 \\
     & 6 & Dual-axis ($\omega=1$) & 52.50 & 61.11 & 48.91 & 54.17 \\
     & 6 & Dual-axis (with $\omega$) & 63.54 & 66.67 & 52.80 & 61.00 \\
\cmidrule(lr){2-7}
     & 4 & Dual-axis ($\omega=1$) & 21.04 & 39.90 & 34.98 & 31.97 \\
     & 4 & Dual-axis (with $\omega$) & 26.67 & 50.51 & 39.51 & 38.89 \\
\bottomrule
\end{tabular}
\end{table}

As shown in Table~\ref{tab:spatial_design_ablation},
dual-axis scaling substantially outperforms both row-only
and column-only scaling on both models. Furthermore,
incorporating row-impact weighting consistently improves
accuracy across all evaluated benchmarks and both bit budgets.
These results validate the effectiveness of both dual-axis
scaling and row-impact weighting in our spatial design.

\paragraph{Lifetime Weighting Ablation.}
We further evaluate the benefit of lifetime-aware allocation
by setting $L_u=1$, corresponding to distortion-only
mixed-precision allocation. We keep the quantizer, bit budget,
FP16 pivots, and allocation algorithm unchanged, and compare
the results using only the temporal component under the
6-bit setting.

\begin{table}[!htbp]
\centering
\caption{Effect of lifetime weighting under the 6-bit
temporal-only configuration with BF16 weights.
All other configurations are kept unchanged.}
\label{tab:lifetime_ablation}
\small
\setlength{\tabcolsep}{6pt}
\begin{tabular}{@{}llrrrr@{}}
\toprule
Model & Method & AIME & GPQA-D & LCB & Avg. \\
\midrule
Qwen & $L_u=1$ & 37.71 & 54.55 & 41.94 & 44.73 \\
     & With $L_u$ & 74.58 & 79.67 & 72.64 & 75.63 \\
\midrule
Kimi & $L_u=1$ & 51.88 & 61.87 & 43.93 & 52.56 \\
     & With $L_u$ & 61.67 & 66.16 & 48.91 & 58.91 \\
\bottomrule
\end{tabular}
\end{table}

As shown in Table~\ref{tab:lifetime_ablation},
removing lifetime weighting consistently degrades performance
across all evaluated benchmarks and both models. The
three-task average drops from 75.63\% to 44.73\% on Qwen
and from 58.91\% to 52.56\% on Kimi. These results
demonstrate that lifetime-aware allocation provides
substantial benefits beyond distortion-based mixed-precision
allocation and FP16 pivot protection.

\subsection{Long-Horizon Error Behavior}
\label{app:long_horizon_error}

We further examine the long-horizon behavior of recurrent-state quantization from two perspectives: prediction drift over extended decoding and the propagation of an isolated quantization error.

\paragraph{Long-horizon prediction drift.}
We extend the forced-decode evaluation to approximately 64K tokens on Qwen3.8-27B and measure excess NLL relative to FP32 across decoding intervals.
From Table~\ref{tab:long_horizon_drift}, 6-bit STEPQuant maintains low excess NLL over long decoding horizons, with a slope of only 0.00018 per 1K tokens compared with 0.058 for uniform INT6. This indicates that STEPQuant substantially mitigates long-horizon error accumulation.

\begin{table}[!htbp]
\centering
\small
\setlength{\tabcolsep}{3.5pt}
\caption{\textbf{Long-horizon prediction drift on Qwen3.8-27B.}
Excess NLL is measured relative to FP32 over approximately 64K forced-decoding tokens.}
\label{tab:long_horizon_drift}
\resizebox{\linewidth}{!}{
\begin{tabular}{@{}lcccccccccc@{}}
\toprule
State & Mean & 0--8K & 8--16K & 16--24K & 24--32K & 32--40K & 40--48K & 48--56K & 56--63K & Slope / 1K \\
\midrule
STEPQuant@6 & 0.0043 & -0.0007 & 0.0014 & 0.0014 & 0.0036 & 0.0052 & 0.0052 & 0.0105 & 0.0085 & 0.00018 \\
INT6        & 2.276  & 0.748   & 1.245  & 1.579  & 2.010  & 2.562  & 2.878  & 3.718  & 3.871  & 0.058 \\
\bottomrule
\end{tabular}
}
\end{table}

\paragraph{Single-error propagation.}
We further measure the propagation of a single injected quantization error over subsequent decoding steps. Values are normalized to the readout error at step 1 and averaged within each interval.
We observe that isolated quantization errors decay substantially over subsequent decoding steps, with the average readout error falling below 5\% of its initial value over steps 129--1024.

\begin{table}[!htbp]
\centering
\small
\setlength{\tabcolsep}{5pt}
\caption{\textbf{Propagation of a single injected quantization error.}
Readout errors are normalized to the error at the first step.}
\label{tab:single_error_propagation}
\begin{tabular}{@{}llcccc@{}}
\toprule
Model & State & Step 1 & Steps 2--16 & Steps 17--128 & Steps 129--1024 \\
\midrule
Qwen3.8-27B         & STEPQuant@6 & 1.000 & 0.501 & 0.152 & 0.0420 \\
Qwen3.8-27B         & STEPQuant@4 & 1.000 & 0.365 & 0.055 & 0.0066 \\
Kimi-Linear-48B-A3B & STEPQuant@6 & 1.000 & 0.273 & 0.036 & 0.0083 \\
Kimi-Linear-48B-A3B & STEPQuant@4 & 1.000 & 0.276 & 0.031 & 0.0026 \\
\bottomrule
\end{tabular}
\end{table}

\subsection{Calibration Robustness}
\label{app:calibration_robustness}

We evaluate the sensitivity of STEPQuant to calibration domain, sample count, and sequence length at 6 bits, and further examine the stability of the resulting precision assignments and FP16 pivot selections.

\paragraph{Calibration sensitivity.}
We measure excess NLL and PPL relative to FP32 over 8K decode steps on four PG-19 books under different calibration configurations.
Across these settings, excess NLL varies by at most 0.012 nats/token, indicating that STEPQuant is robust to the evaluated calibration configurations.

\begin{table}[!htbp]
\centering
\small
\setlength{\tabcolsep}{6pt}
\caption{\textbf{Sensitivity to calibration configurations.}
The reference configuration uses 32 WikiText-2 segments of length 2048.}
\label{tab:calibration_sensitivity}
\begin{tabular}{@{}lcccc@{}}
\toprule
Calibration & Qwen excess NLL & Qwen PPL & Kimi excess NLL & Kimi PPL \\
\midrule
WikiText-2, 32 $\times$ 2048 (reference) & 0.1072 & 9.612 & 0.0248 & 10.068 \\
Github, 32 $\times$ 2048                 & 0.1044 & 9.585 & 0.0131 & 9.951 \\
WikiText-2, 8 $\times$ 2048              & 0.1089 & 9.628 & 0.0261 & 10.081 \\
WikiText-2, 32 $\times$ 512              & 0.1054 & 9.594 & 0.0326 & 10.146 \\
\bottomrule
\end{tabular}
\end{table}

\paragraph{Calibration plan stability.}
We further compare the precision assignments and FP16 pivot selections under different calibration settings with those obtained using the reference configuration (WikiText-2, 32 $\times$ 2048). Here, Agree denotes the fraction of units assigned the same bit width, while Pivot overlap denotes the fraction of shared FP16 pivots.
As shown in Table~\ref{tab:calibration_plan_stability}, bit-width assignments remain highly consistent across calibration settings, with at least 91.3\% agreement on both models. Together with the small variation in excess NLL, these results demonstrate that STEPQuant is robust to changes in calibration domain, sample count, and sequence length.

\begin{table}[!htbp]
\centering
\small
\setlength{\tabcolsep}{5pt}
\caption{\textbf{Stability of precision assignments and FP16 pivot selections across calibration configurations.}}
\label{tab:calibration_plan_stability}
\begin{tabular}{@{}lcccc@{}}
\toprule
Calibration & Qwen agree & Qwen pivot overlap & Kimi agree & Kimi pivot overlap \\
\midrule
WikiText-2, 8 $\times$ 2048  & 0.991 & 1.000 & 0.982 & 0.926 \\
WikiText-2, 16 $\times$ 2048 & 0.992 & 1.000 & 0.984 & 0.941 \\
WikiText-2, 64 $\times$ 2048 & 0.993 & 1.000 & 0.987 & 0.943 \\
WikiText-2, 32 $\times$ 512  & 0.961 & 0.875 & 0.937 & 0.607 \\
WikiText-2, 32 $\times$ 4096 & 0.985 & 1.000 & 0.974 & 0.855 \\
Pile-CC, 32 $\times$ 2048    & 0.956 & 0.875 & 0.945 & 0.711 \\
Github, 32 $\times$ 2048     & 0.924 & 0.938 & 0.913 & 0.568 \\
ArXiv, 32 $\times$ 2048      & 0.946 & 0.938 & 0.929 & 0.736 \\
\bottomrule
\end{tabular}
\end{table}

\section{SGLang Integration, Decode Throughput, and State-Pool Memory}
\label{app:serving}
\subsection{Packed-State Integration in SGLang}
\label{app:backends}
We implement the compressed-state path in SGLang
\citep{zheng2024sglang}, pinned to version 0.5.12, with checkpoint-specific
precision maps, pivot identities, preconditioners, and packed-state kernels.
The adapter maps request slots and tensor-parallel shards onto the shared
recurrent-state representation. Prefill unpacks active states, runs the native chunk kernel,
and packs the resulting boundary states. Decode updates integer pages directly
using floating-point tiles, with no persistent full-matrix FP32 shadow in the
serving path. Slot initialization and reuse follow SGLang's request lifecycle.
As described in Section~\ref{sec:kernel}, reconstruction, the Delta update,
and the current readout are fused. After the readout, scale fitting and
packed writeback run on a separate CUDA stream while later layers process
the token. Writeback completes before the next recurrent step reads the state.
Benchmark evaluations use the SGLang serving path.

For Qwen STEPQuant@6bit, the packed recurrent-state pages occupy
29,999,104 bytes (28.609\,MiB) per request,
including integer codes, FP16 scales, and FP16 pivots, compared with
150,994,944 bytes (144\,MiB) for FP32.
This gives a $5.03\times$ storage reduction (80.13\%).

\subsection{Decode Timing Protocol and Throughput Results}
\label{app:speed_protocol}

We measure decode throughput from consecutive CUDA start events on each
TP rank, including gaps between decode steps. For each run, elapsed decode
time is the sum of those intervals on the slowest rank. Dividing the
number of request-token transitions by this time gives tokens/s.
Prefill, startup, warmup, and final output delivery lie outside this interval.
Each configuration has three runs after one warmup run. We report the
median throughput.

All batches use four A800 GPUs with TP4, $r=1$ (running states only), and the same 128-token prompt
for every request, with 1,024 generated tokens. Decoding is greedy. Every recorded decode step
retains its stated batch size. FP32 state and STEPQuant@6bit share server settings
within each pair. The compressed path includes fitting, FP16 pivots, and
writeback.

\subsection{Memory Reserved for Concurrent Requests}
\label{sec:capacity_motivation}

Let $B$ denote supported concurrent requests, $L$ context length, $W$ shared
text weights, $S$ one aggregate persistent-state slot, and $K$ attention
KV per token. With reservation ratio $r$ and one additional sentinel slot,
\begin{equation}
 M(B,L)=W+(rB+1)S+BLK.
 \label{eq:serving_capacity}
\end{equation}
It should be noted that this model describes reserved serving-memory capacity, rather than measured peak GPU memory usage.
For Qwen, an FP32 slot contains 150,994,944 bytes (144\,MiB) of recurrent
matrices and 2,949,120 bytes (2.8125\,MiB) of convolution state, so
$S=146.8125$\,MiB. Across 16 full-attention layers,
$K=65,536$ bytes (64\,KiB) per context token.

For the radix-caching configuration considered here, SGLang reserves
three slots per supported request. Overlap tracking adds two, plus one global sentinel slot
\citep{zheng2024sglang}. Figure~\ref{fig:state_capacity}a uses this
five-slot configuration, $(5B+1)S$, before and after compression.
Under this five-slot reservation, the state term is independent of context
length.

Weights are counted from the local checkpoints' safetensors payloads,
including integer codes, scales, zero points, and retained BF16 text
weights in the W4A16 model. Vision, MTP draft, and shape metadata are
excluded. The exact text-weight payloads are
53,791,996,928 bytes (BF16) and 17,776,688,128 bytes (W4A16).
All counts are summed across tensor-parallel ranks for one model replica.
The analytical capacity curve reserves 30,015,488 bytes of
compact recurrent-state representation, including pivot scale slots, plus the unchanged
2,949,120-byte convolution state, totaling 31.4375\,MiB. Applying this
format uniformly to the reserved slots gives 9.85\,GiB of reserved
persistent-state pool capacity at $B=64$, compared with 46.02\,GiB in
FP32. Weights and attention KV are excluded from the state curves. Horizontal lines show the two weight payloads.

\subsection{Additional Results on KDA}
Section~\ref{sec:serving_throughput} presents the serving-level
and state-level efficiency results on Qwen.
Here, we provide the corresponding results on KDA
under the same evaluation settings.

\begin{figure}[!htb]
    \centering
    \includegraphics[width=0.7\linewidth]{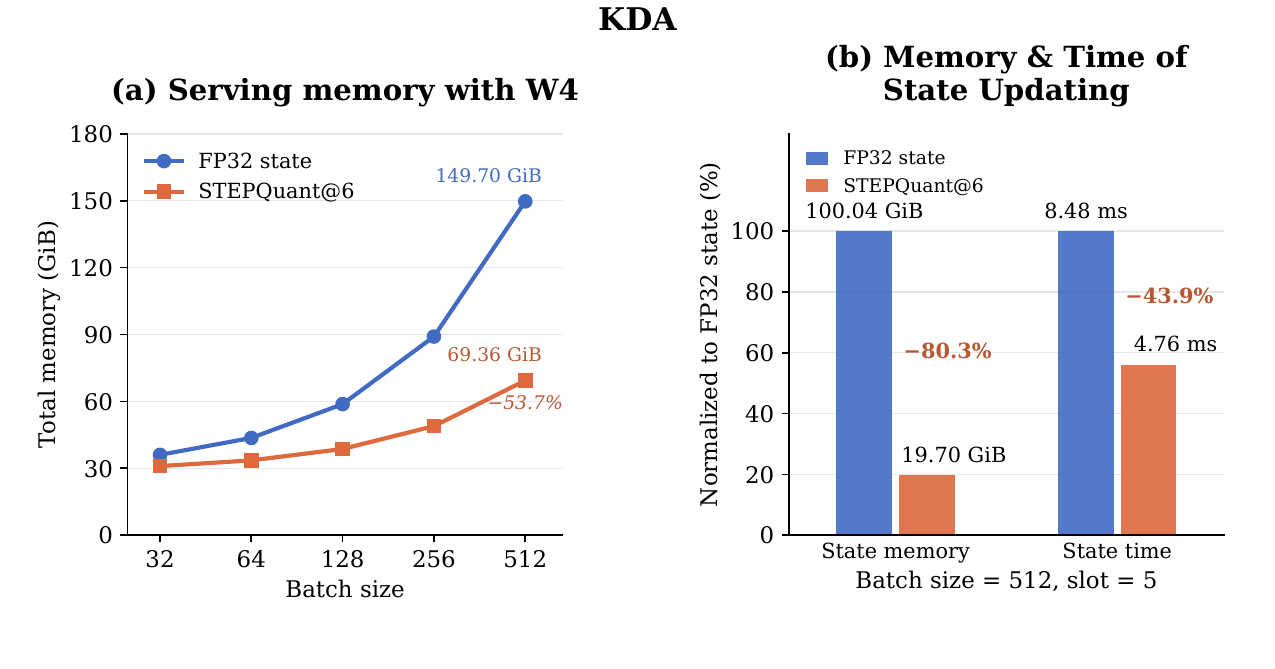}
    \caption{Serving memory and state-update time of STEPQuant on KDA.}
    \label{fig:app_kda_efficiency}
\end{figure}

As shown in Fig.~\ref{fig:app_kda_efficiency},
STEPQuant@6bit reduces total serving memory
from 149.70 to 69.36 GiB (53.7\%) at a batch
size of 512.
At the state level, memory consumption decreases
from 100.04 to 19.70 GiB (80.3\%, 5.08$\times$ compression),
while state-update time decreases from
8.48 to 4.76 ms (43.9\%, 1.78$\times$ faster).
These results further demonstrate the memory
and computational benefits of STEPQuant on KDA.

\subsection{Decode Throughput}
\label{app:throughput}
We evaluate full-model decode throughput with BF16 weights
on four NVIDIA A800 GPUs using tensor parallelism of four
(TP4). We test batch sizes $B\in\{32,64,128,256,512\}$,
with 128 input tokens and 1,024 generated tokens per request.
The measurements include inter-step gaps, quantization
fitting, and compressed-state writeback.

\begin{table}[!htb]
 \centering\small\setlength{\tabcolsep}{7pt}
 \caption{\textbf{Decode throughput.}
 Units are tokens/s. Gains use the unrounded throughputs.}
 \begin{tabular}{lrrrr}
\toprule
Model & Batch & FP32 & STEPQuant@6bit & Gain \\
\midrule
Qwen & 32 & 1,656 & 1,727 & +4.32\% \\
Qwen & 64 & 2,980 & 3,122 & +4.77\% \\
Qwen & 128 & 4,157 & 4,485 & +7.87\% \\
Qwen & 256 & 5,655 & 6,360 & +12.47\% \\
Qwen & 512 & 6,040 & 7,280 & +20.53\% \\
\midrule
KDA & 32 & 5,248 & 5,356 & +2.06\% \\
KDA & 64 & 8,284 & 8,523 & +2.89\% \\
KDA & 128 & 13,284 & 13,936 & +4.90\% \\
KDA & 256 & 16,979 & 18,505 & +8.99\% \\
KDA & 512 & 21,241 & 23,748 & +11.80\% \\
\bottomrule
\end{tabular}

 \label{tab:serving_throughput}
\end{table}

\begin{figure}[!htb]
    \centering
    \includegraphics[width=0.7\linewidth]{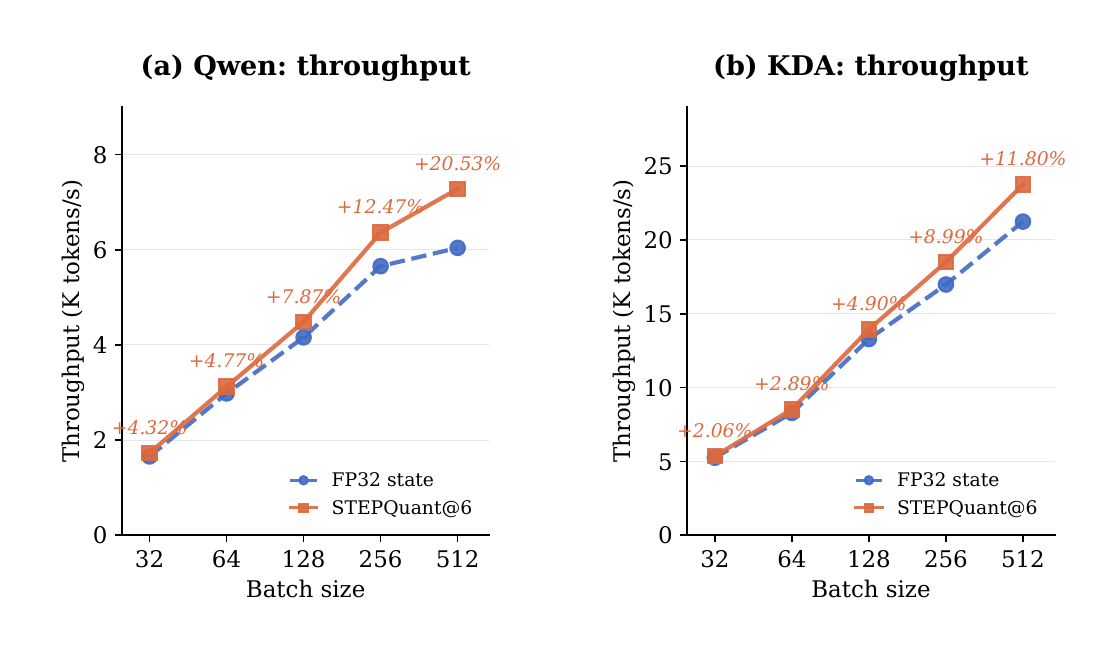}
    \caption{Full-model decode throughput.}
    \label{fig:app_throughput}
\end{figure}

As shown in Fig.~\ref{fig:app_throughput},
STEPQuant@6bit consistently improves decode throughput
on both models, with larger gains at higher batch sizes.
At $B=512$, throughput increases by 20.53\% on Qwen
(from 6,040 to 7,280 tokens/s) and 11.80\% on KDA
(from 21,241 to 23,748 tokens/s).
Detailed results are reported in
Table~\ref{tab:serving_throughput}.

\subsection{Peak Memory Measurement}
\label{app:peak_memory}

We additionally measure peak GPU memory usage under the
four-GPU throughput configuration (BF16 weights, TP4, $r=1$).
STEPQuant@6bit reduces peak GPU memory from 178.40 GiB
to 125.99 GiB, saving 52.41 GiB (29.4\%).
Unlike the reserved-capacity results in Fig.~\ref{fig:serving_throughput}(c),
these measurements reflect actual runtime GPU memory usage,
including additional system overhead.

\FloatBarrier
\section{Comparison with DAMP}
\label{app:damp_comparison}

The concurrent work DAMP~\citep{zhang2026damp} uses reconstruction error and decay-based persistence to select FP16 key channels, while quantizing the remaining channels to INT8 after a Hadamard transform. In contrast, STEPQuant combines lifetime-aware mixed-precision allocation with key-row-aware dual-axis fitting, accounting for both error persistence and readout impact.

Since DAMP has not been open-sourced, we re-implement its original configuration following~\citet{zhang2026damp}, which results in an effective storage cost of 9.88 bits per state value. For a controlled comparison at a comparable 6-bit budget, we further construct a lower-bit variant that retains the same FP16 channel selection and Hadamard transform while quantizing the remaining channels to INT4 instead of INT8. This yields an effective bit width of 6.38 bits, closely matching STEPQuant@6bit (6.36 bits on Qwen3.8-27B and 6.30 bits on Kimi-Linear-48B-A3B-Instruct). Controlled comparisons under identical evaluation settings are reported in Section~\ref{sec:results} (Table~\ref{tab:reasoning}).

In addition, we compare relative FP32 accuracy on three KDA benchmarks shared with DAMP. We directly extract the results of DAMP and its INT8, INT4, and NVFP4 baselines from the original paper and normalize each method's three-task average by its corresponding FP32 baseline. As shown in Table~\ref{tab:damp_comparison}, DAMP retains 100.99\% of FP32 accuracy at 9.9 effective bits per state value, whereas STEPQuant retains 100.51\% at 6.30 bits and 92.91\% at 4.30 bits. The INT8, INT4, and NVFP4 baselines reported by DAMP retain 83.54\%, 6.66\%, and 6.74\%, respectively. While differences in evaluation settings preclude a strictly controlled cross-study comparison, these results further support STEPQuant's favorable accuracy-storage trade-off at low precision.

\begin{table}[!htbp]
\centering
\caption{Accuracy retention on three shared KDA benchmarks:
AIME 2026, HMMT Feb 2026, and LiveCodeBench v6.
Retention is normalized to each study's FP32 baseline.}
\label{tab:damp_comparison}
\setlength{\tabcolsep}{6pt}
\renewcommand{\arraystretch}{0.95}
\begin{tabular}{@{}lcc@{}}
\toprule
Method & Avg. bits & Retention (\%) \\
\midrule
DAMP & 9.9 & 100.99 \\
\rowcolor{methodwash}
STEPQuant@6bit & 6.3 & 100.51 \\
\rowcolor{methodwash}STEPQuant@4bit & 4.3 & 92.91 \\
\midrule
INT8$^{\dagger}$ & 9.0 & 83.54 \\
INT4$^{\dagger}$ & 5.0 & 6.66 \\
NVFP4$^{\dagger}$ & 4.5 & 6.74 \\
\bottomrule
\addlinespace[2pt]
\multicolumn{3}{@{}l}{\footnotesize
$^{\dagger}$ Baseline results reported by DAMP.}
\end{tabular}
\end{table}

\FloatBarrier
\section{Generalization to Mamba2}
\label{app:mamba2}

We further evaluate STEPQuant on Mamba2-2.7B~\citep{dao2024transformers} to examine its generalization beyond GDN and KDA. Mamba2 differs from GDN/KDA in its recurrent update, using only the $\alpha$ decay term and a transposed state formulation. We directly quantize its recurrent state and compare STEPQuant with uniform quantization and Q-Mamba on WikiText-2 perplexity. Table~\ref{tab:mamba2} reports the results together with the exact storage cost in bits per value.

STEPQuant remains effective on Mamba2, closely matching FP32 at 6 bits and retaining strong performance at 4 bits. In contrast, both uniform quantization and Q-Mamba degrade substantially at lower bit widths. These results show that STEPQuant generalizes beyond GDN/KDA to a different recurrent-state formulation.

\begin{table}[!htbp]
\centering
\small
\setlength{\tabcolsep}{6pt}
\caption{\textbf{Recurrent-state quantization on Mamba2-2.7B.}
Exact storage costs include the corresponding quantization metadata.}
\label{tab:mamba2}
\begin{tabular}{@{}lccc@{}}
\toprule
State & Bits/value & WikiText-2 PPL $\downarrow$ & vs. FP32 \\
\midrule
FP32             & 32.00 & 9.062 & -- \\
INT8             & 8.125 & 9.158 & +1.06\% \\
INT6             & 6.125 & 15.22 & +68.0\% \\
INT4             & 4.125 & 272.7 & collapse \\
Q-Mamba@8        & 8.375 & 9.147 & +0.95\% \\
Q-Mamba@6        & 6.375 & 9.526 & +5.13\% \\
Q-Mamba@4        & 4.375 & 14.82 & +63.6\% \\
\rowcolor{methodwash}
\textbf{STEPQuant@6} & \textbf{6.42} & \textbf{9.072} & \textbf{+0.11\%} \\
\rowcolor{methodwash}
\textbf{STEPQuant@4} & \textbf{4.60} & \textbf{9.266} & \textbf{+2.25\%} \\
\bottomrule
\end{tabular}
\end{table}

\FloatBarrier
\section{Limitations}
\label{app:limitations}
While STEPQuant achieves near-lossless accuracy under a 6-bit budget, a modest performance gap remains at 4 bits on certain challenging long-reasoning benchmarks (particularly for sparse MoE models). Achieving strictly lossless compression at 4 bits and pushing recurrent-state quantization to even lower bit widths (e.g., 2--3 bits) remain open directions for future work. In addition, our current kernels primarily target integer formats on Ampere GPUs; extending STEPQuant to emerging low-bit floating-point formats (e.g., NVFP4/MXFP4) on newer hardware architectures is a promising avenue for further exploration.

\end{document}